\documentclass{article}
\usepackage{arxiv}
\usepackage{pifont}
\usepackage{xcolor} 
\usepackage{colortbl} 
\usepackage{multirow}
\usepackage{times}
\usepackage{url}
\usepackage[utf8]{inputenc}
\usepackage{graphicx}
\usepackage{amsmath}
\usepackage{amsthm}
\usepackage{booktabs}
\usepackage{algorithmic}
\usepackage[switch]{lineno}
\usepackage{array}
\usepackage{soul}
\usepackage{caption}
\usepackage{adjustbox}
\usepackage{amssymb}
\usepackage{algorithm}
\usepackage[
    backend=biber,
    style=ieee,
    sorting=count
]{biblatex}
\usepackage[hidelinks]{hyperref}

\newcommand{\eg}{\textit{e}.\textit{g}.}

\title{VMBench: A Benchmark for Perception-Aligned Video Motion Generation}

\author{
Xinran Ling$^{1}$\thanks{Equal Contribution}, Chen Zhu$^{1}$\footnotemark[1], Meiqi Wu$^{1,3}$\footnotemark[1], Hangyu Li$^{1}$, Xiaokun Feng$^{1,2}$, \newline
Cundian Yang$^{1}$, Aiming Hao$^{1}$, Jiashu Zhu$^{1}$, Jiahong Wu$^{1}$\thanks{Corresponding authors}, Xiangxiang Chu$^{1}$
\affiliations
$^1$ AMAP, Alibaba Group, $^2$ CRISE, Institute of Automation, Chinese Academy of Sciences \newline
$^3$ School of Computer Science and Technology, University of Chinese Academy of Sciences
}
\begin{document}

\maketitle

\begin{abstract}
Video generation has advanced rapidly, improving evaluation methods, yet assessing video's motion remains a major challenge. Specifically, there are two key issues: 1) current motion metrics do not fully align with human perceptions; 2) the existing motion prompts are limited. Based on these findings, we introduce \textbf{VMBench}---a comprehensive \textbf{V}ideo \textbf{M}otion \textbf{Bench}mark that has perception-aligned motion metrics and features the most diverse types of motion. 
VMBench has several appealing properties: (1) \textbf{Perception-Driven Motion Evaluation Metrics}, we identify five dimensions based on human perception in motion video assessment and develop fine-grained evaluation metrics, providing deeper insights into models' strengths and weaknesses in motion quality. (2) \textbf{Meta-Guided Motion Prompt Generation}, a structured method that extracts meta-information, generates diverse motion prompts with LLMs, and refines them through human-AI validation, resulting in a multi-level prompt library covering six key dynamic scene dimensions. 
(3) \textbf{Human-Aligned Validation Mechanism}, we provide human preference annotations to validate our benchmarks, with our metrics achieving an average 35.3\% improvement in Spearman's correlation over baseline methods. This is the first time that the quality of motion in videos has been evaluated from the perspective of human perception alignment. Additionally, we will soon release VMBench at \url{https://github.com/GD-AIGC/VMBench}, setting a new standard for evaluating and advancing motion generation models.
\end{abstract}  
\section{Introduction}
\label{sec:intro}

\begin{figure*}
    \centering
    \includegraphics[width=0.9\linewidth]{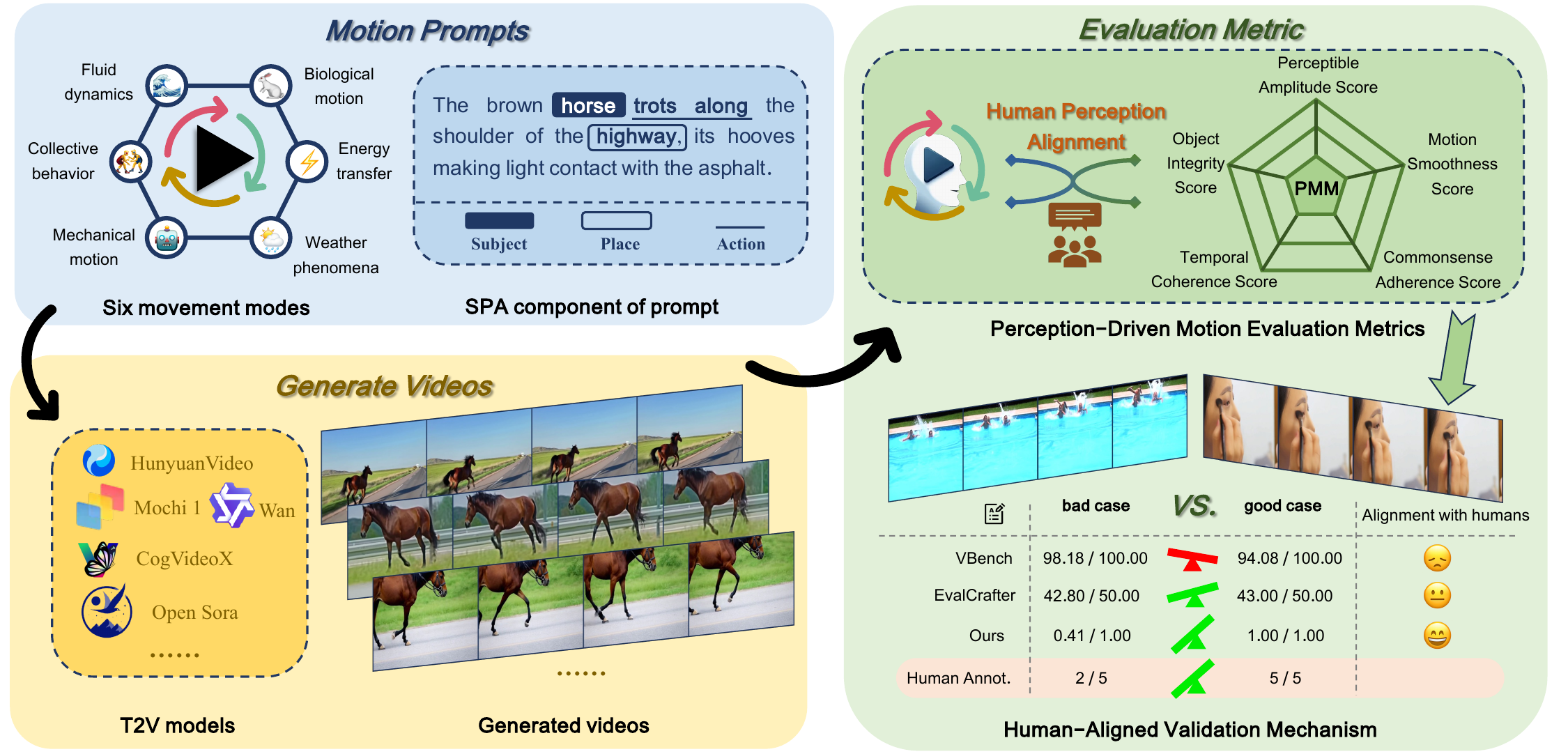}
    \caption{
    Overview of \textbf{VMBench}. Our benchmark encompasses six principal categories of motion patterns, with each prompt constructed as a comprehensive motion structured around three core components: subject, place, and acion. We propose a novel multi-dimensional video motion evaluation comprising five human-centric quality metrics derived from perceptual preferences. Utilizing videos generated by popular T2V models, we conduct systematic human evaluations to validate the effectiveness of our metrics in capturing human perceptual preferences.
    }
    \label{fig:Overview of VMBench.}
\end{figure*}

Video generation models \cite{ni2023conditional, wang2023disco, li2024generative, yang2024cogvideox,kong2024hunyuanvideo,kling,Sora} have advanced significantly, particularly text-to-video (T2V) models \cite{zheng2024opensora,lin2024opensora-plan, genmo2024mochi, blattmann2023stablevideodiffusionscaling}, which perform well in generating static content but struggle to capture motion in alignment with human perception. To enhance dynamic motion generation, it is crucial to establish perception-aligned motion evaluation.

However, previous methods~\cite{heusel2017fid,unterthiner2018fvd,salimans2016isscore,radford2021clip,liu2023fetv} primarily assess the static content of videos, with dynamic evaluation largely limited to motion smoothness~\cite{huang2024vbench,liu2024evalcrafter,liao2025devil}. This makes it challenging to capture motion deficiencies~\cite{zeng2024dawn, sun2024sora}, including but not limited to spatiotemporal inconsistencies and violations of physical laws. \textit{Therefore, developing a more comprehensive evaluation metric that better aligns with human perception is crucial}.

To address this, we introduce Video Motion Benchmark (VMBench) (Fig.\ref{fig:Overview of VMBench.}), the first benchmark focused on comprehensive video motion assessment. It features the most diverse range of motion types, covering 969 categories---more than any other benchmark~\cite{huang2024vbench, liu2024evalcrafter, he2024videoscore, liu2023fetv, sun2024t2vcompbench, liao2025devil} (see Fig.~\ref{fig:prompts statistic1} in the Appendix). Additionally, it integrates perception-aligned motion metrics to systematically evaluate the motion generation quality of video generation models.

First, we introduce Perception-Driven Motion Evaluation Metrics (PMM), the first evaluation framework explicitly designed to align with human perception of motion quality. PMM comprises five key components: Object Integrity Score (OIS), Perceptible Amplitude Score (PAS), Temporal Coherence Score (TCS), Motion Smoothness Score (MSS), and Commonsense Adherence Score (CAS). Unlike previous methods that primarily focus on motion smoothness~\cite{heusel2017fid,unterthiner2018fvd,salimans2016isscore,radford2021clip}, PMM provides a more comprehensive assessment by evaluating motion quality in terms of spatiotemporal inconsistencies and violations of physical laws.

Next, we propose Meta-Guided Motion Prompt Generation (MMPG), an expandable framework that encompasses the most comprehensive range of motion types, grounded in physics \cite{stam2023stable, palomaki2013entangling, rasshofer2011influences, scellier2017equilibrium} and cognitive science \cite{pavlova2012biological, gordon2014ecology}. The evaluation system covers six movement modes: fluid dynamics, biological motion, mechanical motion, weather phenomena, collective behavior, and energy transfer (Fig.~\ref{fig:Overview of VMBench.}). Specifically, MMPG first extracts subjects, places, and actions from VidProm \cite{wang2024vidprom}, Place365 \cite{zhou2017places}, and other datasets~\cite{carreira2019short, bain2021frozen, xu2016msr, anne2017localizing}, then leverages LLMs to enrich metadata and generate detailed motion descriptions. We validate the prompts through human-AI collaboration with DeepSeek-R1 \cite{deepseekai2025deepseekr1incentivizingreasoningcapability} to ensure coherence and rationality. To reduce costs, we curated 1,050 user prompts and use T2V models to generate motion videos for evaluation.

Finally, we provide human preference annotations to validate our benchmarks, showing that our metrics achieve an average 35.3\% improvement in Spearman’s correlation~\cite{zar2005spearman} over current methods \cite{radford2021clip,caron2021dino,huang2024vbench,liu2024evalcrafter,teed2020raft,wu2023dover,wang2025internvideo2.5,bai2025qwenvl,zhang2024llava-video}. This result highlights the enhanced alignment of our evaluation metrics with human perception. We will open-source VMBench, including all prompts, evaluation methods, generated videos, and human preference annotations, and also include more video generation models in VMBench to drive forward the field of video motion generation. Our contributions can be summarized as follows:
\begin{itemize}
    \item First-Ever Human Perception-Aligned Motion Evaluation: We pioneer the evaluation of motion quality in videos from the perspective of human perception alignment.
    \item Meta-Guided Motion Prompt Generation: A curated prompt set designed to assess diverse motion aspects in video models.
    \item Human-Aligned Validation Mechanism: Our metrics achieve an average 35.3\% improvement in Spearman’s correlation over current methods.
\end{itemize}

\section{Related work}
\label{sec:related work}

\begin{figure*}
    \centering
    \includegraphics[width=\linewidth]{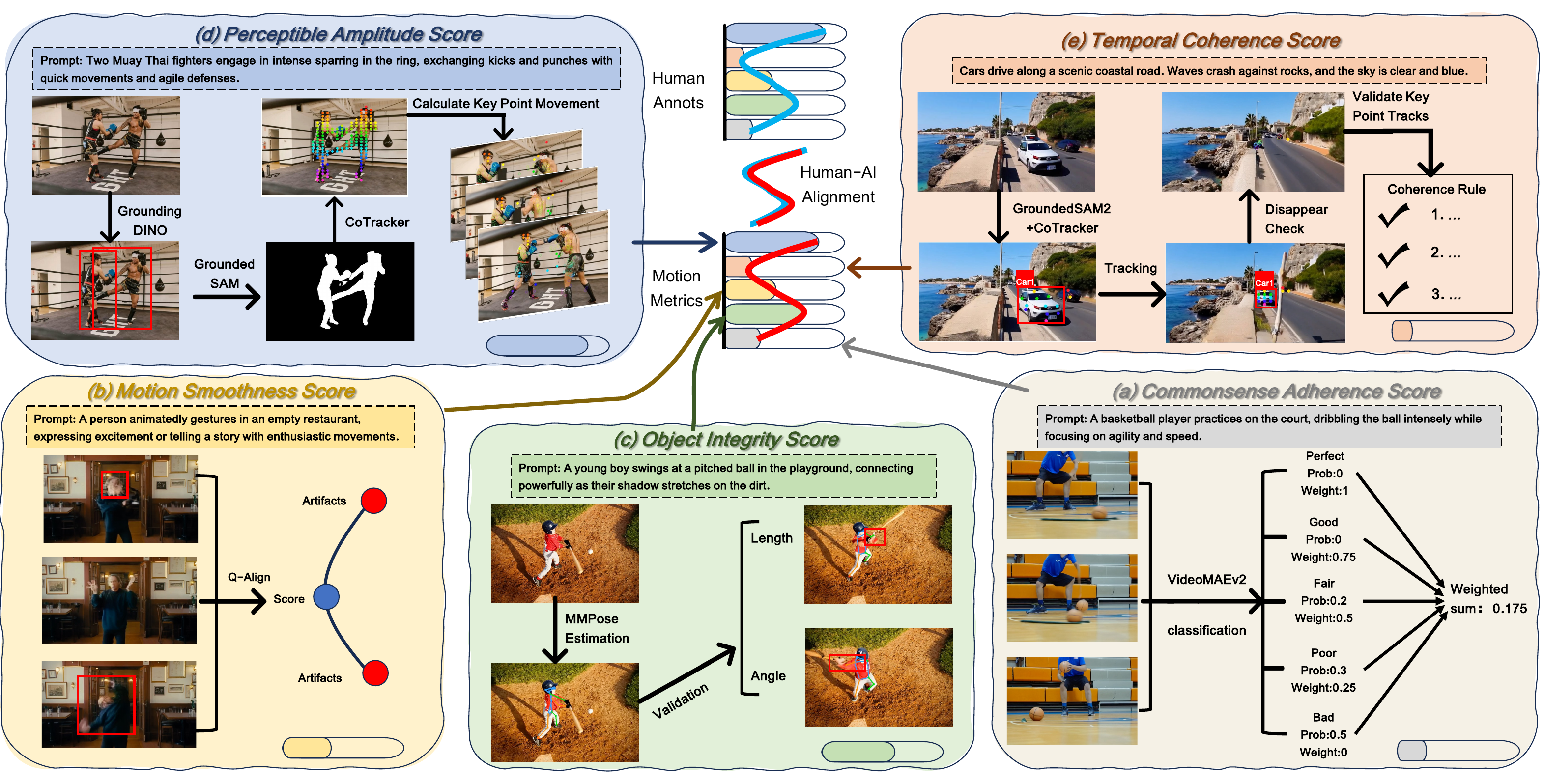}
    \caption{
    Framework of our \textbf{P}erception-Driven \textbf{M}otion \textbf{M}etrics (\textbf{PMM}). PMM comprises multiple evaluation metrics: Commonsense Adherence Score (CAS), Motion Smoothness Score (MSS), Object Integrity Score (OIS), Perceptible Amplitude Score (PAS), and Temporal Coherence Score (TCS). 
    (a-e): Computational flowcharts for each metric. The scores produced by PMM show variation trends consistent with human assessments, indicating strong alignment with human perception.
    }
    \label{fig:evaluation pipeline}
\end{figure*}

\subsection{Video Generative Models}

With the rapid advancement of AI generative techniques----GANs~\cite{li2018video, deng2019irc, kim2020tivgan, gur2020hierarchical, mehmood2024vtm}, VAEs~\cite{li2018video, wu2021godiva}, autoregressive models~\cite{chen2020generative, yu2023magvit, ren2025videoworld}, and diffusion models~\cite{ho2022video, an2023latent, jiang2023text2performer, wang2023modelscopetexttovideotechnicalreport, xing2023dynamicrafter, hong2022cogvideo}---the field of T2V generation has seen remarkable progress. However, while current T2V models~\cite{genmo2024mochi, zheng2024opensora, chen2024videocrafter2, yang2024cogvideox}, excel in generating static visual content, their dynamic performance still deviates significantly from human perception.

Early T2V models~\cite{ho2022video, singer2022make, ho2022imagenvideohighdefinition, chen2023videocrafter1} have built upon Text-to-Image(T2I) techniques, enhancing their temporal capabilities to generate videos. Then, some works~\cite{he2023latentvideodiffusionmodels, zhou2022magicvideo, magicvideov2} take this further by conducting the diffusion process within the latent space and incorporate the spatio-temporal module~\cite{wang2023modelscopetexttovideotechnicalreport, wu2023tuneavideooneshottuningimage, chen2024videocrafter2} within the denoising network, which greatly improves the quality of static content in videos.
CogVideo~\cite{hong2022cogvideo,yang2024cogvideox} and other closed source works~\cite{blattmann2023stablevideodiffusionscaling, polyak2025moviegencastmedia, Gen3, veo2} represented by Sora~\cite{Sora} focus on high-quality data filtering on a large-scale dataset, significantly boosting model performance. OpenSora~\cite{zheng2024opensora} and OpenSora-Plan~\cite{lin2024opensora-plan} aim to showcase and open source the remarkable performance of closed-source models.
Mochi 1~\cite{genmo2024mochi} and HunyuanVideo~\cite{kong2024hunyuanvideo} have improved the dynamic quality of generated videos through new technologies such as asymmetric architecture and full attention mechanism.
Beyond optimization of models and datasets, recent studies~\cite{ma2025step, ling2024motionclone, li2024t2v, veo2, kling} are beginning to explore the influence of motion on video generation. Specifically, ~\textcite{ma2025step} focuses on the motion quality of videos during data selection, in order to enable models to learn motion dynamics even at low resolution training stages. Meanwhile, some works~\cite{ling2024motionclone, li2024t2v} improve the generative capacity of models by incorporating motion information as guidance during the sampling process.

Although these models demonstrate commendable performance across various metrics, the generated videos still exhibit several qualitative motion deficiencies, such as the abrupt disappearance of moving subjects, body distortions, and physically implausible actions, which are difficult to assess using existing metrics.

\subsection{Evaluations Metrics for Video Generation}

Existing approaches for video motion evaluation can be broadly categorized into three paradigms: 1) feature-based metrics leveraging pre-trained video representations, 2) rule-based metrics employing manually designed scoring mechanisms, and 3) MLLM-based assessments that fine-tune large multimodal language models on human-annotated video quality datasets for perceptual scoring.

Previous feature-based metrics demonstrate limited effectiveness in assessing motion quality. Image-centric metrics like IS \cite{salimans2016isscore} and FID \cite{heusel2017fid} inherently ignore temporal coherence, while FVD \cite{unterthiner2018fvd} extends image-level metrics to videos by modeling spatiotemporal features, it reduces motion dynamics to simplistic distributions. These methods conflate feature-space distances with perceptual motion quality, overlooking artifacts like implausible movement.

Recent benchmarks~\cite{huang2024vbench, liu2024evalcrafter} attempt to conduct systematic evaluations using manually curated metrics. For instance, VBench~\cite{huang2024vbench} introduces motion-related criteria such as motion smoothness, quantified via a video frame interpolation model. However, such rule-driven metrics face two inherent limitations in motion assessment: 1) Subjectivity in metric design---predefined rules poorly generalize to diverse motion patterns beyond designer priors; 2) Fragmentary analysis---isolated measurements of individual attributes fail to capture holistic motion quality. Consequently, these heuristics overlook complex phenomena like inertia preservation or multi-object interaction coherence.

Emerging methods \cite{he2024videoscore, bansal2024videophy} employ multimodal large language models (MLLMs) fine-tuned on human preference data to predict perceptual video quality scores. VideoScore \cite{he2024videoscore} aligns scoring with human judgments through preference learning, while VideoPhy \cite{bansal2024videophy} specifically targets physical plausibility via physics-focused annotations. Despite improved alignment with subjective evaluations, these approaches exhibit critical limitations for motion assessment: 1) Oversimplified scoring granularity – scalar outputs obscure specific motion defects; 2) Annotation bias amplification – training data biases toward conspicuous artifacts neglect subtle kinematic violations; 3) Contextual rigidity – domain-specific tuning limits generalization to diverse motion types. While advancing automated evaluation, they struggle to disentangle motion quality from global preference impressions.

Current evaluation paradigms remain fundamentally misaligned with human perception of motion quality due to the above limitations. To bridge this gap, we introduce 5 distinct human-aligned evaluation metrics that systematically quantify motion quality.

\section{VMBench}
\label{sec:VMBench}

Assessing motion in videos continues to present two main issues: current motion metrics do not fully align with human perception, and the range of existing motion prompts is limited, leaving the models' potential in motion generation underexplored. To enhance the motion generation capabilities of video generation models, we introduce VMBench, a comprehensive Video Motion Benchmark that includes perception-aligned motion metrics and features the most diverse range of motion types.

\subsection{Perception-Driven Motion Evaluation Metrics}

When observing videos, the human brain first constructs a holistic understanding of the scene based on prior experiences and physical laws~\cite{Felleman1991DistributedHP}, before selectively attending to moving objects to assess motion smoothness and temporal consistency, particularly when objects are momentarily occluded. While rapid movement can effectively capture attention, excessive deviations from typical motion patterns may induce visual disorientation. As illustrated in Fig.~\ref{fig:metric} in the Appendix, our motion evaluation metrics are designed to emulate this hierarchical perceptual process, transitioning from global scene comprehension to localized motion analysis to ensure coherence and logical consistency.

\noindent\textbf{Commonsense Adherence Score (CAS)} is used to evaluate whether generated videos align with human commonsense. As shown in Fig.~\ref{fig:evaluation pipeline} (a), we train a specialized model capable of assessing the commonsense quality of video content, categorizing it into five levels: Bad, Poor, Fair, Good, and Perfect. First, we collect a dataset comprising 10k generated videos, covering both legacy approaches and polpular models \cite{yang2024cogvideox,kong2024hunyuanvideo,zheng2024opensora,lin2024opensora-plan,chen2023videocrafter1,chen2024videocrafter2,pika,Sora,kling}. Second, we establish perceptual ground truth through systematic pairwise comparisons using VideoReward \cite{liu2025improving}, a reward model trained on 182k human-annotated video preference pairs. This process generates normalized preference scores reflecting human perceptual consensus. Third, we discretize the human preference into five cognitive labels. Using these preference labels, we develop a video classification model to serve as the final CAS predictor, utilizing VideoMAEv2 \cite{wang2023videomae}, a spatiotemporal vision transformer. Compared to MLLMs, this architecture demonstrates temporal modeling capabilities through dense frame sampling and 3D convolution operations. We compute the CAS using a Mean Opinion Score (MOS)~\cite{telecom2000recommendation,wu2024qalign}, where predicte probabilities for each class are weighted by their corresponding quality coefficients: $\text{CAS}=\sum_{i=1}^5p_{i}G(i)$, where $p_i$ denotes the predicted probability for the $i$-th class, and $G_i$ represents the mapping function that converts category to quality weights: $G:i\to w$ (e.g., mapping class 3 to 0.5). For more details, please refer to our Appendix \ref{Commonsense Adherence Score Appendix}.

\noindent\textbf{Motion Smoothness Score (MSS)} aims to detect unsmoothness, like low-level temporal artifacts and high-level motion blur. It addresses the limitations of prior metrics that suffered from either low-level optical flow bias \cite{lai2018warping-error-1,lei2020warping-error-2} or oversimplified motion modeling \cite{li2023amt}, both leading to misalignment with human perception. Considering that natural motion patterns inherently maintain consistent visual quality across temporal sequences, we leverage Q-Align's \cite{wu2024qalign} aesthetic score to detect artifacts, as illustrated in Fig.~\ref{fig:evaluation pipeline} (b). An artifact frame is confirmed when a significant drop in Q-Align scores between consecutive frames exceeds a predefined threshold, mimicking human perception of sustained visual anomalies. Further, our approach employs a scene-aware adaptive thresholding mechanism derived from the statistical modeling of real video segments \cite{carreira2019short,ng2022animal} across diverse motion patterns. The final MSS is computed as: $\text{MSS}=1-\frac{1}{T}\sum_{t=2}^T\mathbb{I}\left(\Delta Q_t>\tau_s(t)\right)$ where $\Delta Q_t$ is the frame-to-frame visual quality degradation magnitude, $\tau_s(t)$ indicates the adaptive threshold and $\mathbb{I}$ is an indicator function. More details are provided in Appendix \ref{Motion Smoothness Score}.

\noindent\textbf{Object Integrity Score (OIS)} detects implausible deformations through spatiotemporal analysis of object integrity. Our approach addresses limitations found in previous methods \cite{caron2021dino}, which primarily focus on object-level semantic consistency. Instead, we focus on detecting perceptual issues (\eg, distorted shapes) that are readily noticeable to the human visual system. As illustrated in Fig.~\ref{fig:evaluation pipeline} (c), we first utilize the MMPose toolkit \cite{mmpose2020} to detect key points of the primary subjects in the generated videos, which are then used to estimate the subjects' shapes in each frame. Then, we quantify shape distortions by analyzing whether object shapes violate real-world anatomical constraints. We establish tolerance thresholds for the above constraints through the statistical analysis of natural motion samples \cite{carreira2019short,ng2022animal,xiang2017posecnn}.
Anatomical violations across consecutive frames will be detected; see more details about anatomical constraints in our appendix \ref{Object Integrity Score Appendix}. The OIS quantifies structural consistency through: $\text{OIS} = \frac{1}{F \cdot K} \sum_{f=1}^{F} \sum_{k=1}^{K} \mathbb{I}\left( \mathcal{D}_f^{(k)} \leq \tau^{(k)} \right)$, in which $F$ and $K$ denote total frame count and anatomical components, respectively. $\mathcal{D}_f^{(k)}$ represents the compound anatomical deviation for component $k$ in frame $f$, and 
$\tau^{(k)}$ indicates statistical thresholds.

\begin{figure*}
    \centering
    \includegraphics[width=\linewidth]{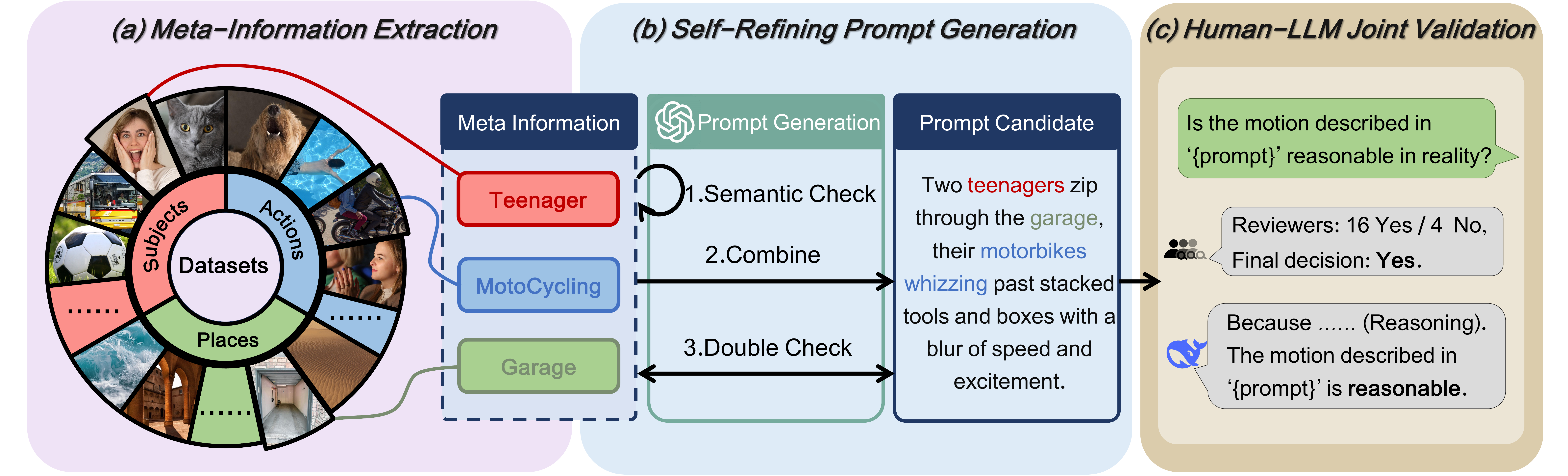}
    \caption{
    Framework of our \textbf{M}eta-\textbf{G}uided \textbf{Motion} \textbf{P}rompt \textbf{G}eneration (\textbf{MMPG}). MMPG consists of three stages:  \textbf{(a) Meta-information Extraction}: Extracting Subjects, Places, and Actions from datasets such as VidProm~\cite{wang2024vidprom}, Didemo~\cite{anne2017localizing}, MSR-VTT~\cite{xu2016msr}, WebVid~\cite{bain2021frozen}, Place365~\cite{zhou2017places}, and Kinect-700~\cite{carreira2019short}. \textbf{(b) Self-Refining Prompt Generation}: Generating and iteratively refining prompts based on the extracted information.  \textbf{(c) Human-LLM Joint Validation}: Validating the prompts through a collaborative process between humans and DeepSeek-R1 to ensure their rationality.}
    \label{fig:pipeline}
\end{figure*}

\noindent\textbf{Perceptible Amplitude Score (PAS)} estimates subject motion by separating it from camera motion, unlike traditional motion magnitude estimation methods~\cite{teed2020raft}, which often overestimate overall motion due to the influence of camera movement. As presented in Fig.~\ref{fig:evaluation pipeline} (d), PAS begins with semantic anchoring via GroundingDINO~\cite{liu2024grounding} to localize moving subjects and distinguish them from passively moving elements. Next, GroundedSAM~\cite{ren2024grounded} ensures temporally stable subject masks, enabling CoTracker~\cite{karaev2024cotracker3} to track key points with high precision. The motion magnitude is then computed based on the average displacement of these key points. Furthermore, PAS accounts for the context-dependent nature of human motion perception, adapting thresholds across different scenarios. To quantify this variability, we derive a set of perceptual motion magnitude thresholds for various scenarios through the statistical analysis of existing video datasets~\cite{carreira2019short,ng2022animal}. These thresholds serve as the foundation for computing a motion score for each video. The PAS is computed as $\text{PAS} = \frac{1}{T} \sum_{t=1}^{T} \min\left( \frac{\bar{D}_t}{\tau_s}, 1 \right)$, where $\bar{D_t}$ denotes frame-level motion amplitude, computed as the average displacement of tracked key points for active subjects in frame $t$ and $\tau_s$ is the perceptual motion threshold for scenario $s$. More Details can be found in appendix~\ref{Perceptible Amplitude Score Appendix}.

\noindent\textbf{Temporal Coherence Score (TCS)} detects the temporal consistency of object motion by analyzing frame-to-frame transitions, addressing the limitation of existing metrics \cite{caron2021dino,radford2021clip} that struggle to differentiate between natural movement and abrupt, unrealistic changes. We combine video object tracking with rule-based validation to detect abnormal disappearance/reappearance patterns, thereby distinguishing natural progression from temporal discontinuities. As shown in Fig.~\ref{fig:evaluation pipeline} (e), first GroundedSAM2 \cite{ravi2024sam2} performs pixel-accurate instance segmentation and tracking across frames, maintaining persistent object IDs throughout the whole sequence. For objects exhibiting discontinuous existence, we apply a secondary verification phase using the CoTracker \cite{karaev2024cotracker3}. It tracks dense key points on target objects and constructs their motion trajectories. We then analyze their motion trajectories to determine whether any anomalous phenomena are present. Our approach mitigates false cases caused by legitimate object discontinuity through a rule-based filtering mechanism implemented via trajectory analysis. These rules account for common scenarios, including 1) objects reappearing after occlusion or disappearing behind obstacles, 2) objects entering or exiting frame boundaries, and 3) apparent size changes due to depth perception, such as objects appearing larger when moving closer or smaller when moving farther away. Through this filtering, we systematically exclude normal events.
The TCS quantifies anomalous discontinuities via $\text{TCS} = 1 - \frac{1}{N} \sum_{i=1}^{N} \mathbb{I}\left( \mathcal{A}_i \land \neg \mathcal{R} \right)$, where $N$ is the total object instances, $\mathcal{A}_i$ indicates the existence status of the $i$-th object, and $\mathcal{R}$ validates legitimate transitions (see Appendix \ref{Temporal Coherence Score Appendix}).

\subsection{Meta-Guided Motion Prompt Generation}

The existing benchmarks~\cite{huang2024vbench, liu2024evalcrafter, he2024videoscore} are constrained by limited and simplistic motion types (As shown in Fig.~\ref{fig:prompts statistic1} of the Appendix), making them inadequate for comprehensively evaluating the motion generation capabilities of video generation models. To address this limitation, we propose an extensible framework---Meta-Guided Motion Prompt Generation (MMPG)---that generates prompts capturing complex motion patterns, as illustrated in Fig.~\ref{fig:pipeline}. Built upon MMPG, we introduce VMBench, a benchmark specifically designed for rigorous evaluation of complex motion generation. We find that VMBench provides the most comprehensive coverage of motion types and the most detailed prompt descriptions, making it an effective benchmark for evaluating the dynamic motion generation capabilities of video generation models. In this section, we elaborate on the process of prompt generation and refinement. More details are available in Appendix~\ref{prompt details}.

\noindent\textbf{Meta-information Extraction.}
We decompose motion descriptions into three key metadata elements: Subject ($S$), Location ($P$) and Action ($A$). We employ Qwen-2.5~\cite{qwen2.5} to extract the metadata library $\{S,P,A\}$ from existing video-text pairs~\cite{wang2024vidprom, xu2016msr, chen2024panda, polyak2025moviegencastmedia}. To ensure novel prompts and enhance textual diversity, we expand the metadata in three key aspects. For subject (\(S\)) classification, we categorize subject types into human, animal, and object, curate a list of principal nouns identifiable by~\cite{ren2024grounded, liu2024grounding}, and use GPT-4o~\cite{achiam2023gpt} to generate descriptions with varying entity counts~(1, 2, n). For place (\(P\)) descriptions, we incorporate data from~\cite{zhou2017places}, filtering out redundant information. For action (\(A\)) expansion, we sample human actions from~\cite{carreira2019short} and use LLM to extend possible actions for animals and objects.

\noindent\textbf{Self-Refining Prompt Generation.} We randomly sample elements from the metadata library $ \{S, P, A \} $ to form metadata sets (\( S^i, P^j, A^k \)) and use GPT-4o to assess their semantic coherence and logical consistency. For coherent sets, we generate prompts of 10 to 60 words that describe the corresponding motion scenarios. To ensure accuracy, we introduce a verification mechanism inspired by~\cite{liu2024evalcrafter}, where the generated prompts are re-evaluated by GPT-4o to verify their consistency with the metadata elements. Through an iterative refinement and filtering process, we generate approximately 50,000 high-quality prompt candidates. However, grammatical correctness alone is insufficient: prompts must also align with physical reality to effectively assess video content. Therefore, we also eliminate descriptions of unrealistic or physically implausible scenarios to ensure the feasibility and authenticity of the prompts.

\noindent\textbf{Human-LLM Joint Validation.} To reduce manual labor and enhance evaluation effectiveness, we further examine and filter prompts through a Human-LLM Joint Validation process. This process combines the strengths of both automated reasoning and human judgment to ensure high-quality motion descriptions. We first leverage the powerful reasoning capabilities of Deepseek R1~\cite{deepseekai2025deepseekr1incentivizingreasoningcapability} to evaluate whether each prompt describes a reasonably realistic motion. After filtering out implausible descriptions, we curate a diverse set of 1,050 prompts with varied metadata. Examples are available in Appendix~\ref{human reasoning validation}.

\section{Experiments}
\label{sec:experiment}

\subsection{Experimental Setup}

\noindent\textbf{Implementation Details.} 
Our benchmark evaluates six popular text-to-video models: OpenSora \cite{zheng2024opensora}, CogVideoX \cite{yang2024cogvideox}, OpenSora-Plan \cite{lin2024opensora-plan}, Mochi 1 \cite{genmo2024mochi}, HunyuanVideo \cite{kong2024hunyuanvideo}, and Wan2.1 \cite{wan2.1}. To provide a richer variety of motion types, we develop the MMPG-set, which comprises 1,050 prompts based on MMPG across six movement modes to assess performance in motion generation. Each model generates 1,050 videos corresponding to the MMPG-set, resulting in a total of 6,300 videos. To ensure a fair comparison, we maintain the hyperparameter settings as defined in each model's project demo. For each prompt, we generate one video for evaluation using only the initial seed. The inference process is executed using 8 Nvidia H20 GPUs. Details of the inference process for each model can be found in Appendix~\ref{inference details}. We then randomly sample 200 videos from each model’s output, resulting in a total of 1,200 videos for human-aligned validation experiments

\noindent\textbf{Comparison Metrics.} The current comparison approaches fall into two categories: Rule-based metrics and Multimodal Large Language Model (MLLM) prompting. (1) Rule-based metrics assess four dimensions, including Perceptible Amplitude, which evaluates through RAFT optical flow magnitude analysis combined with structural motion consistency via 4 frame SSIM averaging \cite{teed2020raft,wang2004ssim}, following established protocols \cite{huang2024vbench}. Temporal Coherence is measured using DINO \cite{caron2021dino} and CLIP \cite{radford2021clip} feature tracking, which relies on cosine similarity between consecutive frames. Motion Smoothness is assessed through a hybrid method that combines interpolation error \cite{li2023amt} with Dover's video quality assessment \cite{wu2023dover}, while Object Integrity is evaluated via dual validation using optical flow warping error \cite{lai2018warping-error-1,lei2020warping-error-2} and semantic consistency checks. 
(2) The MLLM evaluation includes five cutting-edge models: LLaVA-NEXT-Video \cite{zhang2024llava-video}, MiniCPM-V-2.6 \cite{yao2024minicpm}, InternVL2.5 \cite{chen2024internvl}, Qwen2.5-VL \cite{bai2025qwenvl}, and InternVideo2.5 \cite{wang2025internvideo2.5}. These models are evaluated through a standardized assessment process, which involves processing uniformly sampled video frames at 2 frames per second to maintain essential motion patterns while managing computational limits. The evaluation dimensions encompass five aspects: amplitude, coherence, integrity, smoothness, and commonsense adherence, with scoring for each dimension using a 1--5 scale. In particular, to ensure fair comparison, consistent frame sequences and evaluation criteria are maintained across all models. 

\noindent\textbf{Metrics.} 1) \textit{Spearman correlation}~\cite{zar2005spearman}: 
Spearman's rank correlation coefficient, often denoted by \(\rho\), is a measure of the strength and direction of the association between two ranked variables. It assesses how well the relationship between two variables can be described by a monotonic function. Spearman correlation is nonparametric and can effectively evaluate associations in datasets where variables may not follow a normal distribution. Unlike Pearson correlation, which captures linear relationships, Spearman correlation focuses on rank-based associations, making it more robust to outliers and suitable for ordinal data or scenarios with nonlinear dependencies. 2) \textit{Accuracy}: To validate how accurately our metrics align with human preferences, we conduct pairwise comparisons across the 1,200 annotated videos (200 prompts $\times$ 6 models). For each prompt, we evaluate all 15 possible video pairs generated by different models (total 3,000 pairs). Human preference labels are determined by comparing the average expert ratings across all five dimensions (OIS, MSS, CAS, TCS, PAS) – the video with higher aggregated human scores is designated as the ``ground truth" preferred sample. Concurrently, we compute metric preferences by comparing videos’ aggregated PMM scores under identical criteria. Alignment accuracy is quantified as the percentage of pairs where PMM preferences match human judgments, with ties excluded to prioritize unambiguous decisions.

\begin{table}[t]
\definecolor{myyellow}{HTML}{FDE992}
\setlength{\tabcolsep}{1mm}
\centering
\resizebox{\linewidth}{!}{
\begin{tabular}{l|c|ccccc}
\toprule
\multirow{2}{*}{\textbf{Method}} & \textbf{Avg.} & \textbf{CAS} & \textbf{MSS} & \textbf{OIS} & \textbf{PAS} & \textbf{TCS}  \\
& $\rho(\uparrow)$ & $\rho(\uparrow)$ & $\rho(\uparrow)$ & $\rho(\uparrow)$ & $\rho(\uparrow)$ & $\rho(\uparrow)$ \\ \midrule
\multicolumn{7}{l}{\textit{Rule-based}} \\ \midrule
SSIM \cite{wang2004ssim} & 1.6 & -0.9 & -12.1 & 8.3 & \cellcolor{myyellow}17.8 & -4.8 \\
$\text{RAFT}^{*}$ \cite{teed2020raft} & -1.7 & -0.7 & -17.0 & -16.6 & \cellcolor{myyellow}47.7 & -21.9 \\
$\text{CLIP}^*$ \cite{radford2021clip} & 15.0 & 21.5 & 36.5 & 31.7 & -42.7 & \cellcolor{myyellow}28.0 \\
$\text{DINO}^*$ \cite{caron2021dino} & 9.4 & 11.0 & 27.2 & \cellcolor{myyellow}27.4 & -45.8 & \cellcolor{myyellow}27.4 \\
$\text{Dover Technical}^{\dagger}$ \cite{wu2023dover} & 20.6 & 40.2 & 32.6 & \cellcolor{myyellow}34.5 & -6.2 & 2.2 \\
$\text{AMT}^*$ \cite{huang2024vbench} & 3.6 & 2.8 & \cellcolor{myyellow}18.1 & 19.6 & -40.4 & 17.8 \\
$\text{Warping Error}^{\dagger}$ \cite{liu2024evalcrafter} & -6.0 & -11.4 & \cellcolor{myyellow}-19.1 & -9.0 & 33.0 & -23.6 \\ \midrule
\multicolumn{7}{l}{\textit{MLLM Prompting}} \\ \midrule
LLaVa-NEXT-Video \cite{zhang2024llava-video} & 17.1 & 15.2 & 11.0 & 16.1 & 30.8 & 12.4 \\
MiniCPM-V-2.6 \cite{yao2024minicpm} & 16.2 & 6.8 & 17.5 & -2.2 & 27.3 & 31.6 \\
IntenVL2.5 \cite{chen2024internvl} & 19.3 & 18.7 & 20.5 & 14.1 & 42.5 & 1.1 \\
Qwen2.5-VL \cite{bai2025qwenvl} & 15.3 & 24.7 & 19.8 & 3.7 & 14.2 & 14.0 \\
InternVideo2.5 \cite{wang2025internvideo2.5} & 26.9 & 22.7 & 21.9 & 29.6 & 44.3 & 15.8  \\ \midrule
\multicolumn{7}{l}{\textit{Model-Rule Synergy}} \\ \midrule
\textbf{PPM (Ours)} & \textbf{62.2} & \textbf{69.9} & \textbf{77.1} & \textbf{65.8} & \textbf{65.2} & \textbf{54.5} \\ \bottomrule
\end{tabular}
}
\caption{Correlation Analysis Between Evaluation Metrics and Human Scores via Spearman Correlation Coefficient ($\rho \times 100$). 
Superscript $*$ and $\dagger$ denote implementations following VBench \cite{huang2024vbench} and EvalCrafter \cite{liu2024evalcrafter} respectively. 
Yellow backgrounds in Rule-based indicate specific dimension baseline.}
\label{tab:correlation-comparison}
\end{table}

\subsection{Comparison with State-of-the-Art}

\noindent\textbf{Human-aligned Validation Mechanism.} We invite three domain experts to independently annotate each sample based on PMM (Perceptible Amplitude, Temporal Coherence, Object Integrity, Motion Smoothness, and Commonsense Adherence), resulting in 6,000 detailed ratings with high interannotator agreement. Finally, we obtain Likert scores~\cite{nemoto2014likert} from humans for each video across the five dimensions included in PMM. More details can be found in Appendix \ref{sup:annotation}. We assess the alignment between metrics and human perception by calculating the Spearman correlation~\cite{zar2005spearman} between metric scores and expert ratings. A higher Spearman correlation signifies a stronger alignment with human perception.  
 
\noindent\textbf{Comparison with Alternative Metrics.} As shown in Table~\ref{tab:correlation-comparison}, in the MSS evaluation, even advanced metrics such as AMT~\cite{li2023amt} (18.1\%) and Warping Error~\cite{lei2020warping-error-2} (-19.1\%) demonstrate limited discriminative power and produce counterintuitive results, particularly in the context of complex deformations. The OIS evaluation also highlights similar shortcomings, with DINO~\cite{caron2021dino} at 27.4\% and Dover~\cite{wu2023dover} at 34.5\%, both of which fail to capture human sensitivity to structural preservation during motion. In relation to the evaluation of PAS, the SSIM~\cite{wang2004ssim} and RAFT~\cite{teed2020raft} metric based on rules show an alignment efficacy of only 17.8\% and 47.7\%, respectively. In contrast, our approach achieves a remarkable 65.2\% alignment efficacy. For the TCS evaluation, the rule-based metrics CLIP~\cite{radford2021clip} and DINO~\cite{caron2021dino} achieve only 28.0\% and 27.4\% alignment efficacy, respectively, and fail to capture human tolerance for minor inconsistencies while maintaining physical plausibility. In comparison, our method achieves an impressive 54.5\% alignment efficacy. VBench~\cite{huang2024vbench} includes RAFT, CLIP, DINO, and AMT, while EvalCrafter~\cite{liu2024evalcrafter} incorporates Dover Technical and Warping Error. According to the data presented in the table, \textit{\textbf{overall, compared to ours, the motion evaluations of VBench~\cite{huang2024vbench} and EvalCrafter~\cite{liu2024evalcrafter} reveal significantly lower correlations with human perception}}. MLLMs exhibit partial competence in evaluations of the Physical Adequacy Score (PAS) (\eg, InternVideo2.5: 44.3\%) \textit{\textbf{Despite their general capabilities, MLLMs average only a correlation of 10.0\%--30.0\% in all dimensions, highlighting a fundamental misalignment with human perception in evaluating motion quality}}.

\subsection{Ablation Study}

\begin{table}[t]
\setlength{\tabcolsep}{2mm} 
\centering
\resizebox{0.9\linewidth}{!}{
\begin{tabular}{ccccc|c}
\toprule
\textbf{CAS} & \textbf{MSS} & \textbf{OIS} & \textbf{PAS} & \textbf{TCS} & \textbf{Accuracy (\%) $\uparrow$} \\
\midrule
\multicolumn{6}{l}{\textit{Removal-based ablation}} \\
\midrule
\ding{51} & \ding{51} & \ding{51} & \ding{51} & \ding{51} & \textbf{70.6} \\
\ding{51} & \ding{51} & \ding{51} & \ding{51} & \ding{55} & 66.9 \\
\ding{51} & \ding{51} & \ding{51} & \ding{55} & \ding{51} & 68.7 \\
\ding{51} & \ding{51} & \ding{55} & \ding{51} & \ding{51} & 64.6 \\
\ding{51} & \ding{55} & \ding{51} & \ding{51} & \ding{51} & 65.2 \\
\ding{55} & \ding{51} & \ding{51} & \ding{51} & \ding{51} & 64.1 \\
\midrule
\multicolumn{6}{l}{\textit{Addition-based ablation}} \\
\midrule
\ding{55} & \ding{55} & \ding{55} & \ding{55} & \ding{55} & -  \\
\ding{51} & \ding{55} & \ding{55} & \ding{55} & \ding{55} & 58.9 \\
\ding{51} & \ding{51} & \ding{55} & \ding{55} & \ding{55} & 66.1 \\
\ding{51} & \ding{51} & \ding{51} & \ding{55} & \ding{55} & 67.3 \\
\bottomrule
\end{tabular}
}
\caption{Ablation Study on our Motion Metrics. 
Prediction accuracy (\%) of different metric combinations against human preferences is calculated.
\textit{Removal-based ablation} show the effect of individually ablating each metric, while in the \textit{addition-based ablation}, we progressively add each metric to observe its impact.}
\label{tab:ablation}
\end{table}

\noindent\textbf{Ablation Study of Motion Metrics.} According to Table \ref{tab:ablation}, removing any single metric from the evaluation of the basic variants results in a significant decline in accuracy, which highlights the importance of each dimension's evaluation metric within the overall framework. Notably, the removal of the CAS metric leads to the most pronounced performance drop, with accuracy falling to 64.1\%, exceeding the impact of other single-dimension ablations. This demonstrates the critical role of the CAS metric in assessing video quality, aligning closely with the key factors prioritized by humans when perceiving video quality. For the performance-oriented variants, we emulate the human perceptual information processing flow by gradually adding evaluation metrics (refer to Fig.~\ref{fig:metric} in the Appendix), with each additional metric significantly enhancing overall accuracy. This not only demonstrates the effectiveness of incrementally increasing evaluation dimensions but also further verifies the consistency of our evaluation approach with human perceptual mechanisms.

\subsection{Qualitative Analysis}
\begin{figure}[t]
    \centering
    \includegraphics[width=\linewidth]{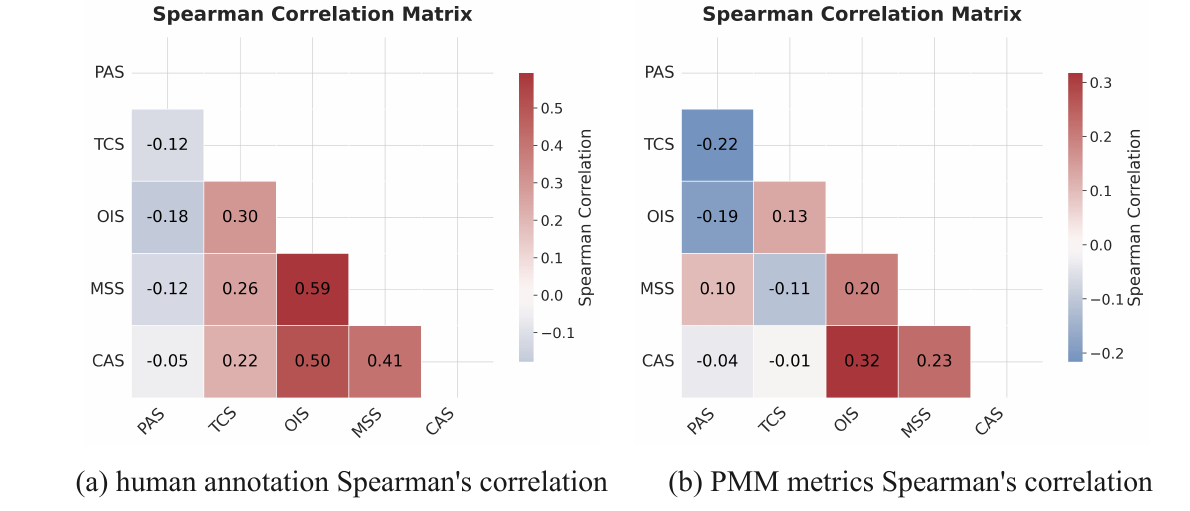}
    \caption{
    Correlation Matrix Analysis of Metrics Within Different Evaluation Mechanisms.
    (a): Spearman Correlation Matrices for human annotations; (b): Spearman Correlation Matrices for our PMM metrics.}
    \label{fig:ablation}
\end{figure}

\noindent\textbf{Alignment of PMM with Human Perception.} As shown in Fig.~\ref{fig:ablation}, the correlation between human scores across these five evaluation dimensions and our evaluation metrics in the same five dimensions remains consistent. For example, both demonstrate a strong correlation between OIS, CAS, and MSS, as well as a weak correlation between PAS and the other metrics. Specifically, as shown in Fig. \ref{fig:ablation} (a), PAS has a negative correlation with other dimensions, such as the Object Integrity Score (OIS), where $\rho = -0.18$. This might be because high dynamic amplitudes in videos cause distortions and artifacts, which lower structural and temporal coherence scores. Conversely, OIS shows strong positive correlations with MSS ($\rho = 0.59$) and CAS ($\rho = 0.50$), suggesting that it reflects physical plausibility and motion rationality well. TCS shows low correlation with other dimensions, which can provide a more comprehensive assessment perspective. PAS and structural/temporal metrics exhibit a negative correlation, which challenges conventional optical-flow-based evaluation frameworks. Additionally, the isolation of PAS highlights the importance of separately disentangling motion magnitude in motion video evaluation. Fig.~\ref{fig:ablation} (b) shows the correlations among the proposed evaluation metrics, which align with human perception.

\begin{table}
\setlength{\tabcolsep}{1mm} 
\centering
\footnotesize 
\resizebox{\linewidth}{!}{%
\begin{tabular}{l|c|ccccc}
\toprule
\textbf{Models} & \textbf{Avg.(\%) $\uparrow$} & \textbf{CAS} & \textbf{MSS} & \textbf{OIS} & \textbf{PAS} & \textbf{TCS} \\
\midrule
Mochi 1\cite{genmo2024mochi} & 53.2 & 37.7 & 62.0 & 68.6 & 14.4 & 83.6  \\
OpenSora \cite{zheng2024opensora}& 51.6 & 31.2 & 61.9 & 73.0 & 3.4 & 88.5  \\
CogVideoX \cite{yang2024cogvideox}& 60.6 & 50.6 & 61.6 & 75.4 & 24.6 & 91.0 \\
OpenSora-Plan \cite{lin2024opensora-plan}& 58.9 & 39.3 & 76.0 & \textbf{78.6} & 6.0 & 94.7 \\
HunyuanVideo \cite{kong2024hunyuanvideo}& 63.4 & 51.9 & 81.6 & 65.8 & \textbf{26.1} & 96.3  \\
Wan2.1 \cite{wan2.1} & \textbf{78.4} & \textbf{62.8} & \textbf{84.2} & 66.0 & 17.9 & \textbf{97.8} \\
\bottomrule
\end{tabular}}
\caption{Performance of Video Generation Models on VMBench. We evaluated six open-source video generate models~\cite{genmo2024mochi, zheng2024opensora, yang2024cogvideox, lin2024opensora-plan, kong2024hunyuanvideo, wan2.1} using VMBench.
A higher score indicates better performance for a category.}
\label{tab:quantitative-analysis}
\end{table}

\noindent\textbf{Assessing Video Generation Models with PMM.} As shown in Table \ref{tab:quantitative-analysis}, we evaluate several leading video generation models using our PMM metrics, including Mochi 1\cite{genmo2024mochi}, OpenSora \cite{zheng2024opensora}, CogVideoX \cite{yang2024cogvideox}, OpenSora-Plan \cite{lin2024opensora-plan}, HunyuanVideo \cite{kong2024hunyuanvideo}, and Wan2.1 \cite{wan2.1}. Our findings indicate that Wan2.1 delivers the best performance in generating motion videos, producing results that appear more realistic. For a clearer visual comparison of the dynamic videos generated by each model, please refer to Fig.~\ref{fig:qualiative-analysis} in the Appendix.

\section{Conclusion}
\label{sec:conclusion}

Video motion authenticity remains a critical challenge in the creation of generate content. In response, we introduce VMBench, the first open-source benchmark for motion quality evaluation, integrating motion metrics with human-aligned assessment to reveal deficiencies in existing models' ability to generate physically plausible movements. To support the research community, we offer three essential resources: 1) a standardized framework for detecting motion artifacts overlooked by traditional metrics; 2) actionable diagnostic tools aimed at guiding model optimization; and 3) quantitative standards that align technical advancements with human perceptual expectations. By establishing a common framework for development in this field, VMBench enables systematic tracking of motion generation, encourages targeted model improvements, and ultimately supports the creation of videos that achieve both visual fidelity and dynamic realism. However, while the benchmark's evaluation metrics are aligned with general human perception, they may not fully capture subtle differences in perception that stem from individual viewer experiences and preferences. Addressing these limitations presents opportunities for further refinement, aiming to ensure a more comprehensive and adaptable approach to video quality evaluation.

\printbibliography
\clearpage
\appendix
\setcounter{page}{1}

\begin{figure*}
    \centering
    \includegraphics[width=0.9\linewidth]{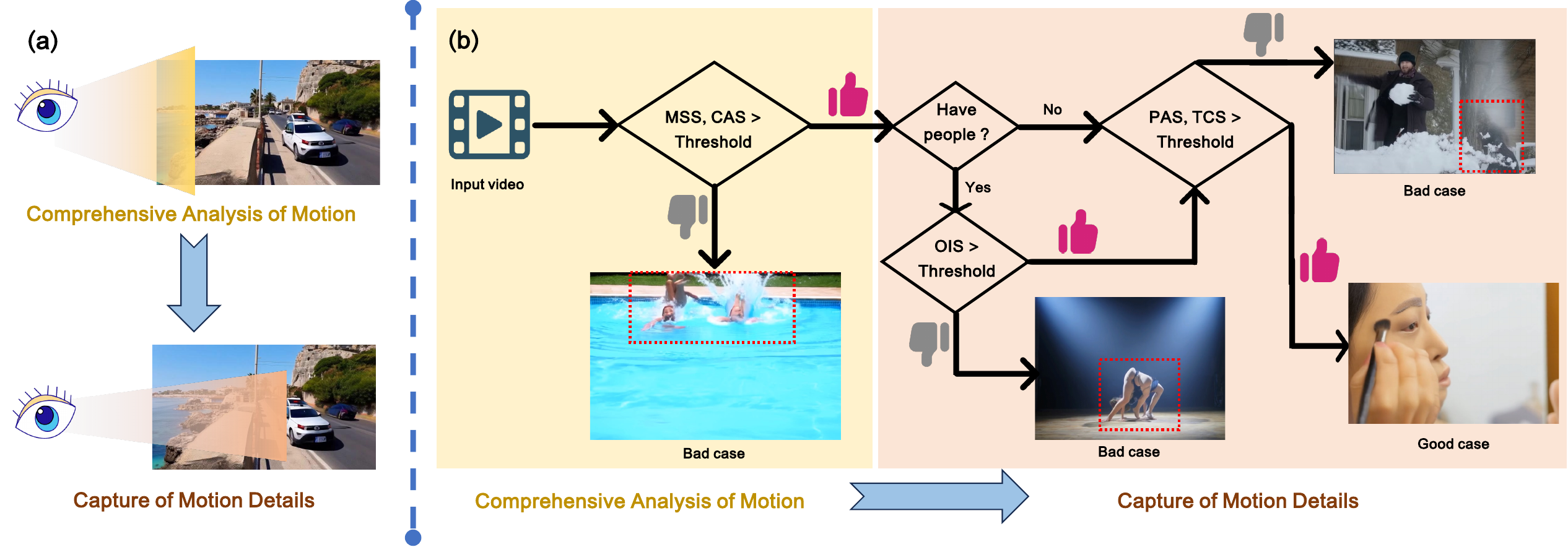}
    \caption{Our metrics framework for evaluating video motion, which is inspired by the mechanisms of human perception of motion in videos.
    (a) Human perception of motion in videos primarily encompasses two dimensions: Comprehensive Analysis of Motion and Capture of Motion Details.
    (b) Our proposed metrics framework for evaluating video motion. Specifically, the MSS and CAS correspond to the human process of Comprehensive Analysis of Motion, while the OIS, PAS, and TCS correspond to the capture of motion details.} 
    \label{fig:metric}
    \hspace{-10mm}
\end{figure*}

\section{Human Perception Flow}
\label{sup:flow}

Driven by the insights of the neuroscientific studies of motion perception, the human perception of motion within video can be systematically decomposed into two primary dimensions at a coarse level: the global parsing of motion fields and the capture of its finer details, as shown in Fig.~\ref{fig:metric}. Specifically, the global perception of video motion fields, facilitates rapid evaluation of generated scenes plausibility by tracking macro-scale motion patterns, such as how smooth motion requires high frame rates to suppress temporal fragmentation (\eg, jitter artifacts). Simultaneously, the fine-grained capture of motion in videos enables the detection of physically implausible movement patterns that violate fundamental physical laws, such as acceleration profiles violating Newton’s laws or trajectories with positional discontinuities. The proposed VMBench systematically decomposes the two fundamental axes into granular perceptual criteria, thereby constructing a multi-dimension evaluation framework to quantitatively assess the spatiotemporal fidelity and motion coherence of generated videos.

\section{Evaluation Dimension}
\subsection{Commonsense Adherence Score (CAS)}
\label{Commonsense Adherence Score Appendix}

\begin{figure*}[ht]
    \centering
    \includegraphics[width=0.9\linewidth]{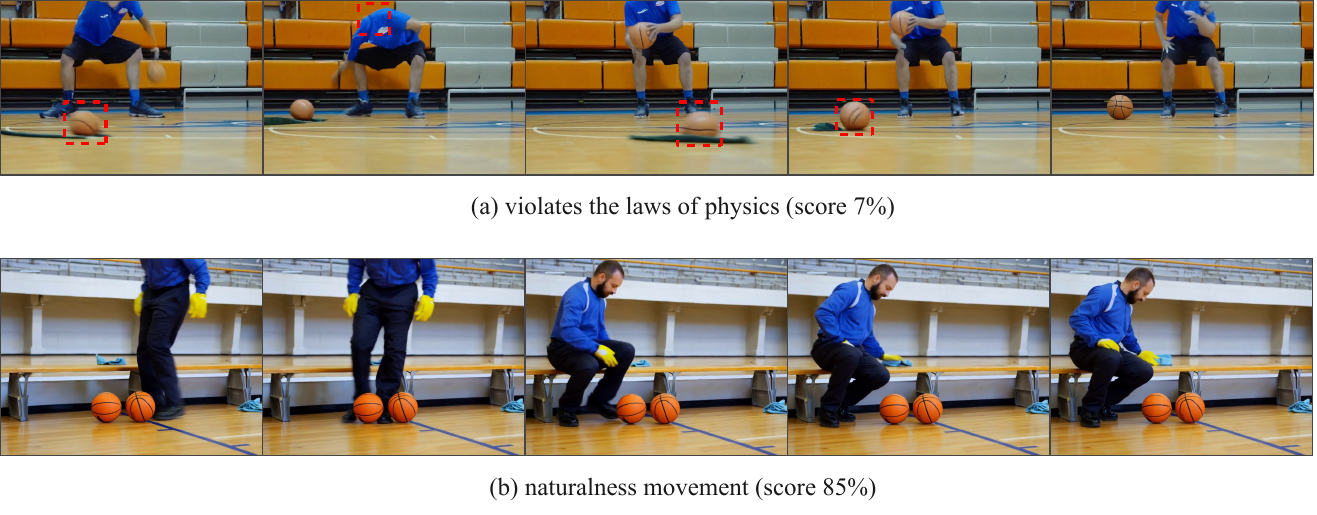}
    \caption{Visualization of Commonsense Adherence. (a) The ball exhibits perpetual rolling motion on the ground without external forces, violating physical laws and contradicting human perception. (b) All objects demonstrate motion consistent with natural physical principles.}
    \label{fig:commonsen-adherence}
\end{figure*}

A prevalent issue in generated videos is the phenomenon that contradicts human perception and physical laws. As demonstrated in Fig.~\ref{fig:commonsen-adherence}, generated videos frequently exhibit motions that defy physical laws and violate everyday intuitions and expectations, significantly compromising realism.
Our CAS aims to evaluate whether generated videos align with human commonsense. As mentioned in the main text, we develop a specialized model to assess the commonsense quality of video content, categorizing it into five levels: Bad, Poor, Fair, Good, and Perfect. 

First, we collect a comprehensive dataset of 10k generated videos from a wide range of sources. This dataset includes videos from legacy approaches as well as those generated by popular models~\cite{yang2024cogvideox, kong2024hunyuanvideo, zheng2024opensora, lin2024opensora-plan, chen2023videocrafter1, chen2024videocrafter2, pika, Sora, kling}. The videos in our dataset come from two main sources: existing web datasets \cite{liu2025improving} and videos that we generate using these models. This approach ensures a diverse representation of video generation techniques and potential outcomes, capturing a wide spectrum of quality levels and possible commonsense violations. Such a comprehensive collection is crucial for training a robust evaluation model capable of assessing various aspects of video quality and realism. Second, we establish perceptual ground truth using VideoReward~\cite{liu2025improving} to conduct systematic pairwise comparisons among the 10k videos. For each video pair, VideoReward determines which is preferable based on human perception standards. We then calculate a win rate for each video, representing its performance in all comparisons. These win rates are used to rank the videos, which are subsequently divided into five equal groups. Each group receives a label indicating its level of adherence to human commonsense expectations, from lowest to highest. Third, we choose the VideoMAEv2~\cite{wang2023videomae} architecture for its temporal modeling capabilities, which are crucial for assessing commonsense adherence in video content. This model processes the input video and outputs logits for each of the five quality categories. We train VideoMAEv2 using the preference labels derived from the previous step. The model is initialized with a ViT-Giant~\cite{dosovitskiy2020vit} backbone pre-trained on large-scale video datasets. We fine-tune this model on our labeled dataset using 8 NVIDIA H20 GPUs.  
Our training process uses a batch size of 10, with input videos resized to $224\times224 pixels$. Each video clip consists of 16 frames, sampled at a rate of 4. We employ the AdamW optimizer with a learning rate of 1e-3 and weight decay of 0.1. The training schedule includes a 5-epoch warm-up period, followed by a total of 35 epochs. To enhance model performance, we implement layer-wise learning rate decay with a factor of 0.9 and a drop path rate of 0.3.

To compute the final CAS, we use a Mean Opinion Score (MOS) approach. The predicted probabilities for each class are weighted by their corresponding quality coefficients. The mapping function $G(i)$ converts the category index to quality weights as follows: $G(1) = 0 (\text{Bad}), G(2) = 0.25 (\text{Poor}), G(3) = 0.5 (\text{Fair}), G(4) = 0.75 (\text{Good}), \text{and }G(5) = 1 (\text{Perfect}).$
The CAS is then calculated using the formula provided in the main text:
\begin{equation}
\text{CAS} = \sum_{i=1}^5 p_i G(i)
\end{equation}
where $p_i$ denotes the predicted probability for the $i$-th class. The resulting score provides a comprehensive measure of how well a generated video aligns with human expectations and commonsense understanding of the world.

\subsection{Motion Smoothness Score (MSS)}
\label{Motion Smoothness Score}

\begin{figure*}[ht]
    \centering
    \includegraphics[width=0.9\linewidth]{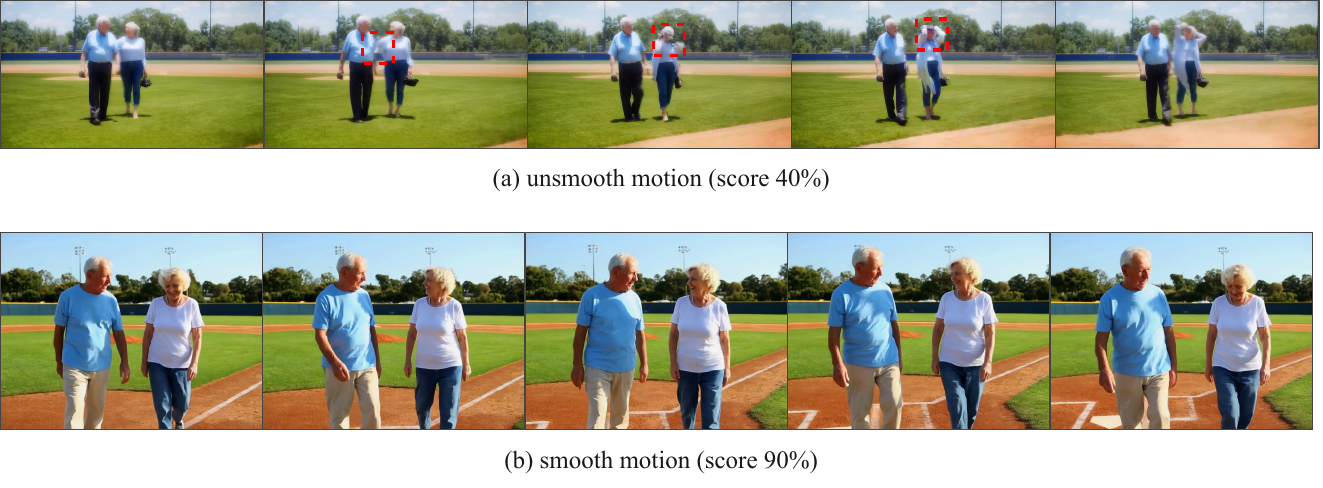}
    \caption{Visualization of Motion Smoothness. (a) Both subjects exhibit significant blur during walking, with the female's facial features particularly affected, resulting in a loss of fine details. (b) Both subjects demonstrate fluid motion, with clear visibility of bodily details.}
    \label{fig:motion-smoothness}
\end{figure*}

Generated videos often exhibit blur and artifacts during object motion, particularly in areas with intricate details. This issue is especially pronounced when depicting complex movements that occur in the real world, as illustrated in Fig.~\ref{fig:motion-smoothness}. These visual inconsistencies likely stem from the model's difficulty in balancing the preservation of fine details with the representation of high-motion changes.

As mentioned in the main text, our MSS leverages Q-Align's \cite{wu2024qalign} aesthetic score to detect artifacts. Here, we provide more details on how we quantify the frame-to-frame visual quality degradation magnitude $\Delta Q_t$. The frame-to-frame visual quality degradation magnitude $\Delta Q_t$ is defined as:
\begin{equation}
\Delta Q_t = Q(f_{t-1}) - Q(f_t)
\end{equation}
where $Q(f_t)$ represents the Q-Align aesthetic score for frame $t$. This formulation captures the change in visual quality between consecutive frames, with positive values indicating a decrease in quality.  
To determine the adaptive threshold $\tau_s(t)$, we conduct a statistical analysis of real video segments from datasets such as \cite{carreira2019short} and \cite{ng2022animal}. We analyze the relationship between motion amplitude and acceptable levels of quality degradation across diverse motion patterns. The threshold $\tau_s(t)$ allows for a higher tolerance of quality degradation in scenes with more intense motion. By incorporating this adaptive thresholding mechanism, our MSS effectively accounts for varying levels of acceptable blur in different motion scenarios, providing a more perceptually aligned evaluation of motion smoothness in generated videos. The final MSS is computed as:
\begin{equation}
\text{MSS} = 1 - \frac{1}{T}\sum_{t=2}^T \mathbb{I}(\Delta Q_t > \tau_s(t))
\end{equation}
The MSS ranges from 0 to 1, where a score of 1 indicates perfect motion smoothness (no frames with significant quality drops), and lower scores indicate a higher proportion of frames with noticeable artifacts or blur.

\subsection{Object Integrity Score (OIS)}
\label{Object Integrity Score Appendix}

\begin{figure*}[ht]
    \centering
    \includegraphics[width=0.9\linewidth]{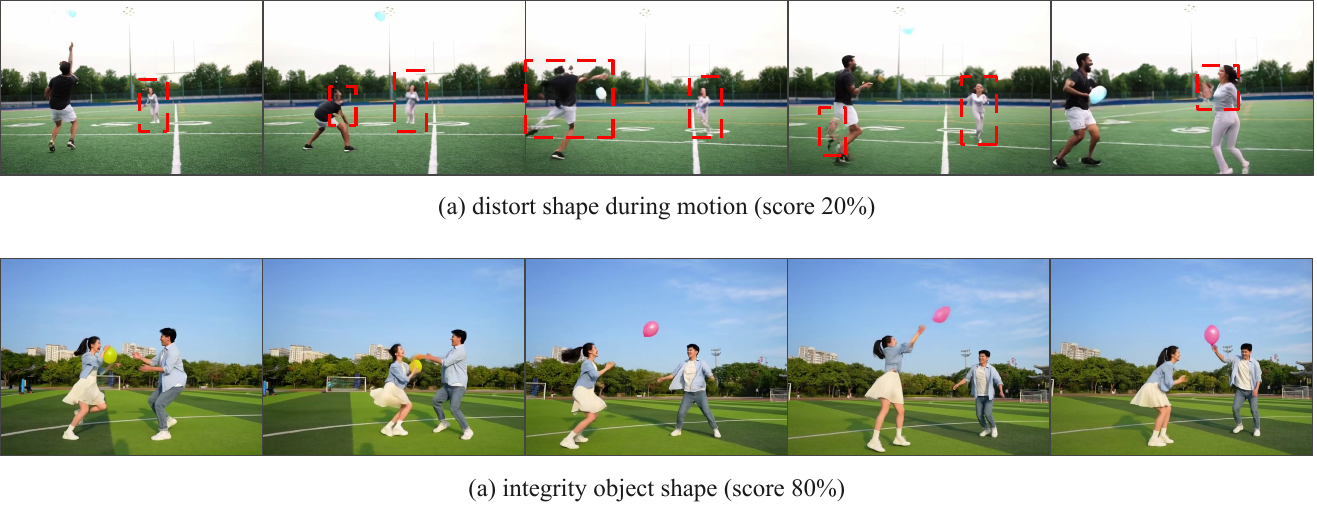}
    \caption{Visualization of Object Integrity. (a) Both subjects exhibit varying degrees of bodily distortion, with their limbs becoming difficult to discern due to severe warping. (b) Both subjects maintain normal anatomical structure throughout the sequence, displaying no unnatural deformations.}
    \label{fig:object-integrity}
\end{figure*}

\begin{figure*}[ht]
    \centering
    \includegraphics[width=0.9\linewidth]{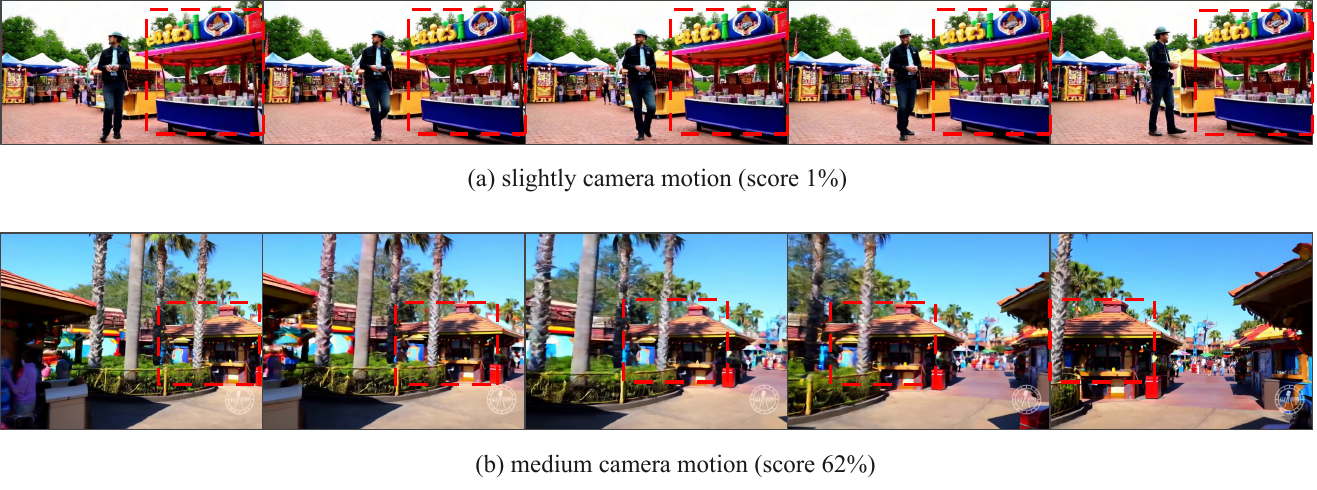}
    \caption{
    Visualization of Camera Motion. (a) The object and background remain relatively static, indicating subtle camera movement. (b) The scene exhibits noticeable changes, demonstrating a panning or tracking camera movement.
    }
    \label{fig:camera-motion}
\end{figure*}

The integrity of moving objects in the generated videos is a crucial factor affecting the overall quality. Object integrity refers to the degree to which objects in the video maintain their physical structure and appearance consistent with real-world expectations. As illustrated in Figure \ref{fig:object-integrity}, generated videos can sometimes exhibit abnormal distortions or deformations of moving objects. These distortions violate our perceptual expectations of normal object behavior and movement. We employ the MMPose toolkit \cite{mmpose2020} to detect key points of the primary subjects in the generated videos. These key points are then used to estimate the subjects' shapes in each frame. Our focus is on detecting perceptual issues (\eg, distorted shapes) that are readily noticeable to the human visual system.

For a comprehensive anatomical analysis, we consider both length and angle variations of object components. Let $K = {k_1, k_2, ..., k_n}$ be the set of key points detected in each frame. Through statistical analysis of our datasets, we establish thresholds $\tau_L$ and $\tau_\theta$ to detect changes in unnatural shape in lengths and angles, respectively.

For length analysis, we calculate the Euclidean distance $L_{i,j}(t)$ between connected key points $k_i$ and $k_j$ in each frame $t$. We then observe the variations in these lengths across frames, identifying potential distortions when changes exceed the threshold $\tau_L$:
\begin{equation}
D_L(i,j) = \sum_{t=2}^T \mathbb{I}(|L_{i,j}(t) - L_{i,j}(t-1)| > \tau_L)
\end{equation}
where $D_L(i,j)$ denotes the distortion count for the component between keypoints $k_i$ and $k_j$, $T$ represents the total number of frames, and $\mathbb{I}(\cdot)$ is the indicator function.

Similarly, for angle analysis, we compute the angles $\theta_{i,j,k}(t)$ formed by adjacent key points in each frame. We monitor these angles for abrupt changes that surpass the threshold $\tau_\theta$:
\begin{equation}
D_\theta(i,j,k) = \sum_{t=2}^T \mathbb{I}(|\theta_{i,j,k}(t) - \theta_{i,j,k}(t-1)| > \tau_\theta)
\end{equation}
These length and angle analyses contribute to the compound anatomical deviation $\mathcal{D}_f^{(k)}$ for each anatomical component $k$ in frame $f$. We establish tolerance thresholds $\tau^{(k)}$ for each anatomical component through statistical analysis of natural motion samples from datasets such as \cite{carreira2019short,ng2022animal,xiang2017posecnn}.

The OIS is then computed as:
\begin{equation}
\text{OIS} = \frac{1}{F \cdot K} \sum_{f=1}^{F} \sum_{k=1}^{K} \mathbb{I}\left( \mathcal{D}_f^{(k)} \leq \tau^{(k)} \right)
\end{equation}
This formulation checks if the compound anatomical deviation $\mathcal{D}_f^{(k)}$ is within the acceptable threshold $\tau^{(k)}$ for each frame and anatomical component. The indicator function returns 1 for each instance where the deviation is within the threshold. We sum these values across all frames and anatomical components and then normalize by dividing by the total number of checks performed $(F \cdot K)$.

\subsection{Perceptible Amplitude Score (PAS)}
\label{Perceptible Amplitude Score Appendix}

\begin{figure*}[ht]
    \centering
    \includegraphics[width=0.9\linewidth]{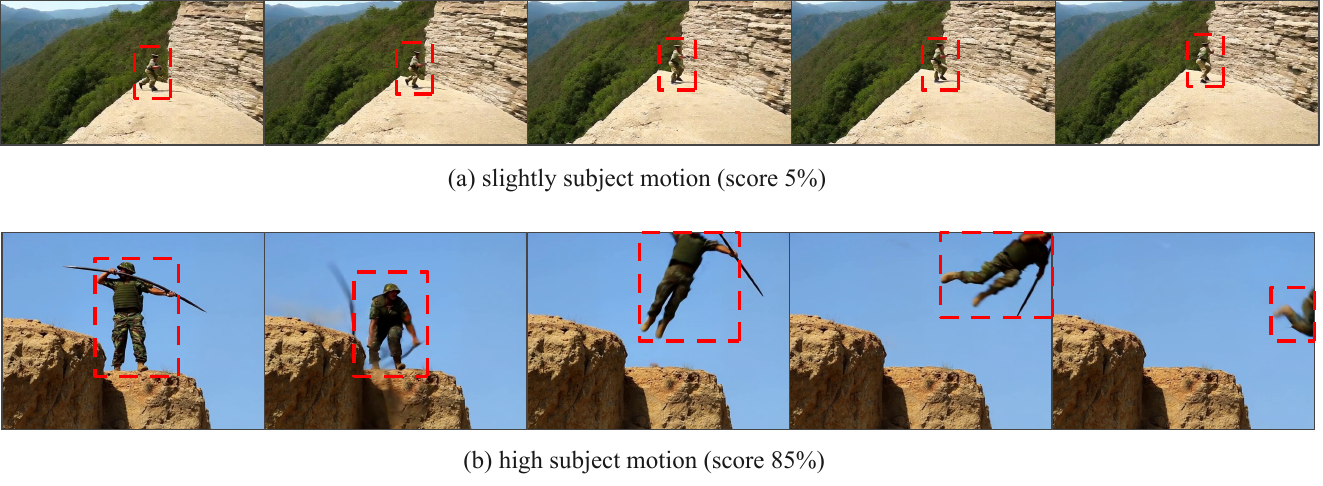}
    \caption{Visualization of Subject Motion. (a) The main subject exhibits only minor changes throughout the video, indicating limited movement. (b) The subject completes a full range of actions, even moving out of frame, demonstrating a significant magnitude of movement.}
    \label{fig:subject-motion}
\end{figure*}

Motion amplitude in videos stems from two sources: camera motion, as illustrated in Fig. \ref{fig:camera-motion}, and subject motion, as demonstrated in Fig. \ref{fig:subject-motion}. Our PAS focuses on the latter. Traditional methods like RAFT \cite{teed2020raft} can be affected by camera motion when detecting subject movement. However, our approach effectively isolates subject motion from camera movement, enabling a more accurate perception of the primary subject's motion regardless of camera dynamics.

Our method begins by employing GroundingDINO \cite{liu2024grounding} to detect the primary moving subject in the video, followed by GroundedSAM \cite{ren2024grounded} to generate precise masks for this subject across frames. We then utilize CoTracker \cite{karaev2024cotracker3} to track key points for the main subject using these masks.

The motion magnitude is computed based on the average displacement of these key points. For each tracked key point $p$ at frame $t$, we calculate its displacement as:
\begin{equation}
D(p^t) = \sqrt{(x_t - x_{t-1})^2 + (y_t - y_{t-1})^2}
\end{equation}
The frame-level motion amplitude $\bar{D_t}$ is then calculated as the average displacement across all tracked key points for active subjects in frame t:
\begin{equation}
\bar{D}_t = \frac{1}{N_t} \sum_{i=1}^{N_t} D(p_i^t)
\end{equation}
where $N_t$ is the number of tracked key points in frame t.
To account for the context-dependent nature of human motion perception, we derive a set of perceptual motion magnitude thresholds $\tau_s$ for various scenarios $s$ through statistical analysis of existing video datasets \cite{carreira2019short,ng2022animal}. These thresholds serve as the foundation for computing a motion score for each video.
The Perceptible Amplitude Score (PAS) is then computed as:
\begin{equation}
\text{PAS} = \frac{1}{T} \sum_{t=1}^{T} \min\left( \frac{\bar{D}_t}{\tau_s}, 1 \right)
\end{equation}
where $T$ is the total number of frames in the video, $\bar{D_t}$ is the frame-level motion amplitude, and $\tau_s$ is the perceptual motion threshold for scenario s.
This method ensures that the PAS accounts for both the magnitude of motion and its perceptual significance in different contexts, providing a more nuanced evaluation of motion in videos.

\begin{figure*}[ht]
    \centering
    \includegraphics[width=0.9\linewidth]{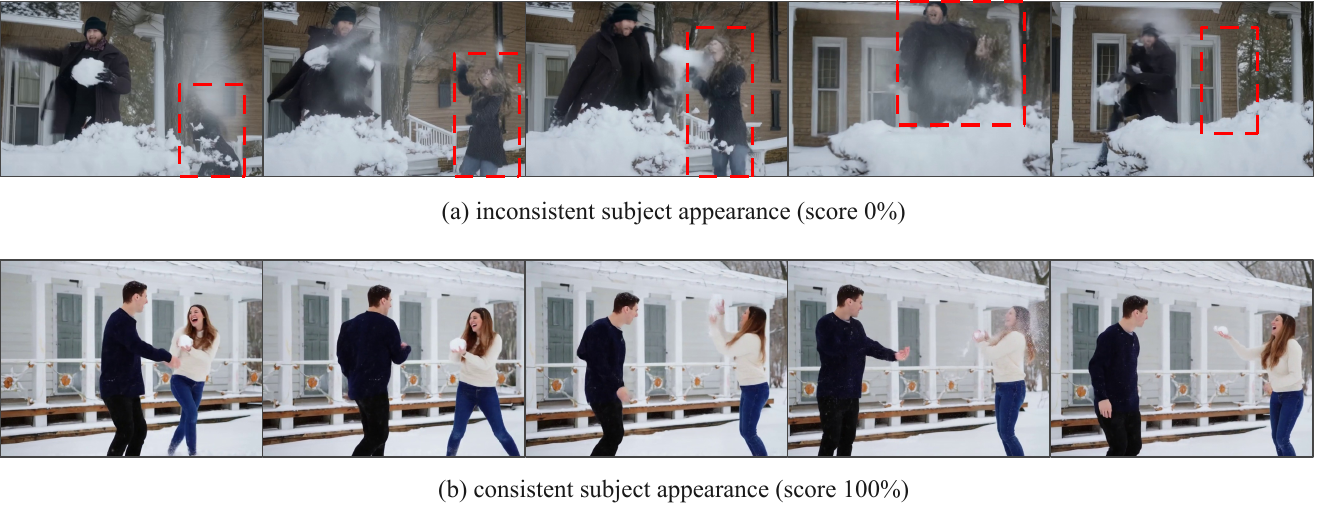}
    \caption{Visualization of Temporal Coherence. (a) The female disappears and reappears throughout the video, while the male exhibits discontinuous behavior. (b) Both subjects maintain consistent presence and stability throughout the sequence, demonstrating superior temporal continuity.}
    \label{fig:temporal-consistency}
\end{figure*}

\subsection{Temporal Coherence Score (TCS)}
\label{Temporal Coherence Score Appendix}
In generated video sequences, moving subjects often exhibit phenomena of sudden disappearance or appearance, as illustrated in Fig. \ref{fig:temporal-consistency}. These temporal discontinuities significantly impact the perceived quality of motion. Stable temporal coherence is crucial for achieving high-quality motion in generated videos.

We employ GroundedSAM2 \cite{ravi2024sam2} for pixel-accurate instance segmentation and tracking across frames, maintaining persistent object IDs throughout the whole sequence. For objects exhibiting discontinuous existence, we apply a secondary verification phase using CoTracker \cite{karaev2024cotracker3} to track dense key points on target objects and construct their motion trajectories.

We then analyze these motion trajectories to determine whether any anomalous phenomena are present. Our approach mitigates false cases caused by legitimate object discontinuity through a rule-based filtering mechanism. These rules account for common scenarios, including: 1) Objects reappearing after occlusion or disappearing behind obstacles. 2) Objects entering or exiting frame boundaries. 3) Apparent size changes due to depth perception, such as objects appearing larger when moving closer or smaller when moving farther away.
Let $N$ be the total number of object instances in the video. For each object instance $i$, we define: $\mathcal{A}$: An indicator function that equals 1 if the object exhibits discontinuous existence, and 0 otherwise. $\mathcal{R}$: A function that validates legitimate transitions based on our rule-based filtering mechanism. It returns 1 if the transition is legitimate (i.e., falls under one of the three scenarios mentioned above), and 0 otherwise.
The TCS is then computed as:
\begin{equation}
\text{TCS} = 1 - \frac{1}{N} \sum_{i=1}^{N} \mathbb{I}(\mathcal{A}_i \land \neg \mathcal{R})
\end{equation}
where $\mathbb{I}(\cdot)$ is the indicator function that returns 1 if the condition inside the parentheses is true, and 0 otherwise. The term $\mathcal{A}_i \land \neg \mathcal{R}$ identifies objects that exhibit discontinuous existence $(\mathcal{A}_i=1)$ and do not have a legitimate reason for this discontinuity $\mathcal{R}=0$.
TCS ranges from 0 to 1, where a score of 1 indicates perfect temporal coherence (no anomalous discontinuities), and lower scores indicate a higher proportion of unjustified object vanishing or emerging events.
This formulation ensures that the TCS accounts for both the presence of discontinuities and the legitimacy of these discontinuities based on our rule-based filtering, providing a nuanced evaluation of temporal coherence in videos.

\begin{figure*}[ht]
    \centering
    \includegraphics[width=0.9\linewidth]{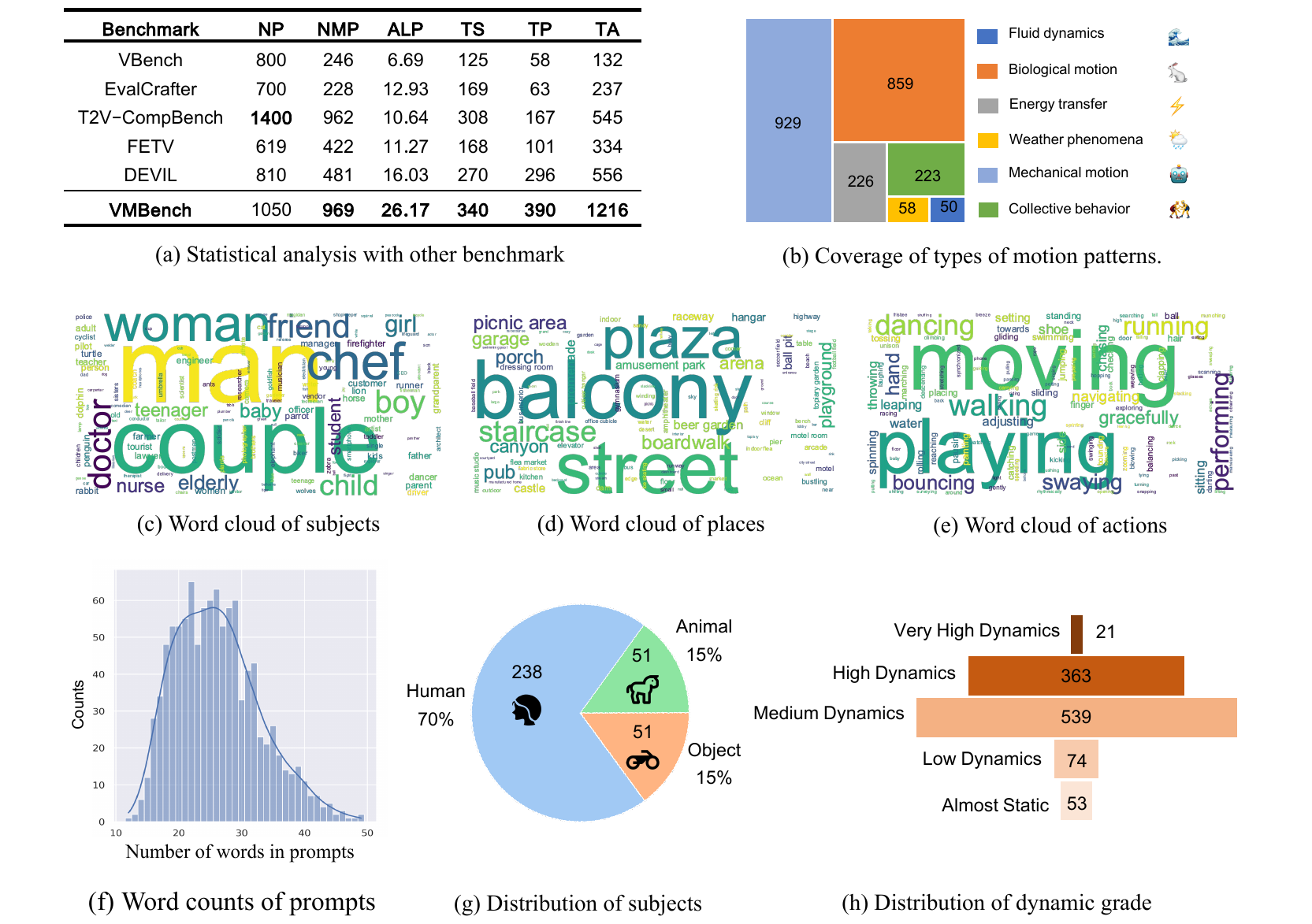}
    \caption{
    Statistical analysis of motion prompts in VMBench. 
    (a-h): Multi-perspective statistical analysis of prompts in VMBench. 
    These analyses demonstrate VMBench's comprehensive evaluation scope, encompassing motion dynamics, information diversity, and real-world commonsense adherence.
    }
    \label{fig:prompts statistic1}
\end{figure*}

\section{MMPG}
\subsection{Prompts Statistic}
\label{prompt details}
In this section, we conduct Motion Prompts Statistics~(as shown in Fig.~\ref{fig:prompts statistic1}) to emphasize VMBench's focus on motion. In Table~(a), we perform a statistical analysis to demonstrate the superiority of our prompts compared to previous works, focusing on the number of prompts~(NP), the number of motion prompts~(NMP), the average length of prompts~(ALP), the types of motion subjects~(TS), place~(TP), and actions~(TA). We find that VMBench provides the most comprehensive coverage of action types and the most detailed prompt descriptions, making it an effective benchmark for evaluating the dynamic motion generation capabilities of video generation models. Fig.~\ref{fig:prompts statistic1} (b) illustrates the distribution pattern of our motion prompts. It is evident that our prompts, while covering six major motion patterns, are particularly rich in content related to the most common mechanical and biological motions found in everyday life. This aligns with the characteristic of our prompts being realistic and sensible descriptions.
Fig.~\ref{fig:prompts statistic1} (c), Fig.~\ref{fig:prompts statistic1} (d), and Fig.~\ref{fig:prompts statistic1}(e) respectively demonstrate the richness of subjects, places and actions within the prompts, highlighting the richness and variety of motion content.
Fig.~\ref{fig:prompts statistic1} (f) presents a well-distributed range of prompt lengths, and Fig.~\ref{fig:prompts statistic1} (g) shows the distribution of motion subjects, reflecting the diversity among subjects in our prompts. 
We employ the dynamic evaluation method from DEVIL~\cite{liao2025devil} to assess the dynamic grade of our prompts, as shown in Fig.~\ref{fig:prompts statistic1} (h). The results indicate that our prompts exhibit a high level of dynamism overall, which poses a challenge for large models.

\subsection{Human-LLM Reasoning Validation}
\label{human reasoning validation}

\begin{figure*}
    \centering
    \includegraphics[width=0.9\linewidth]{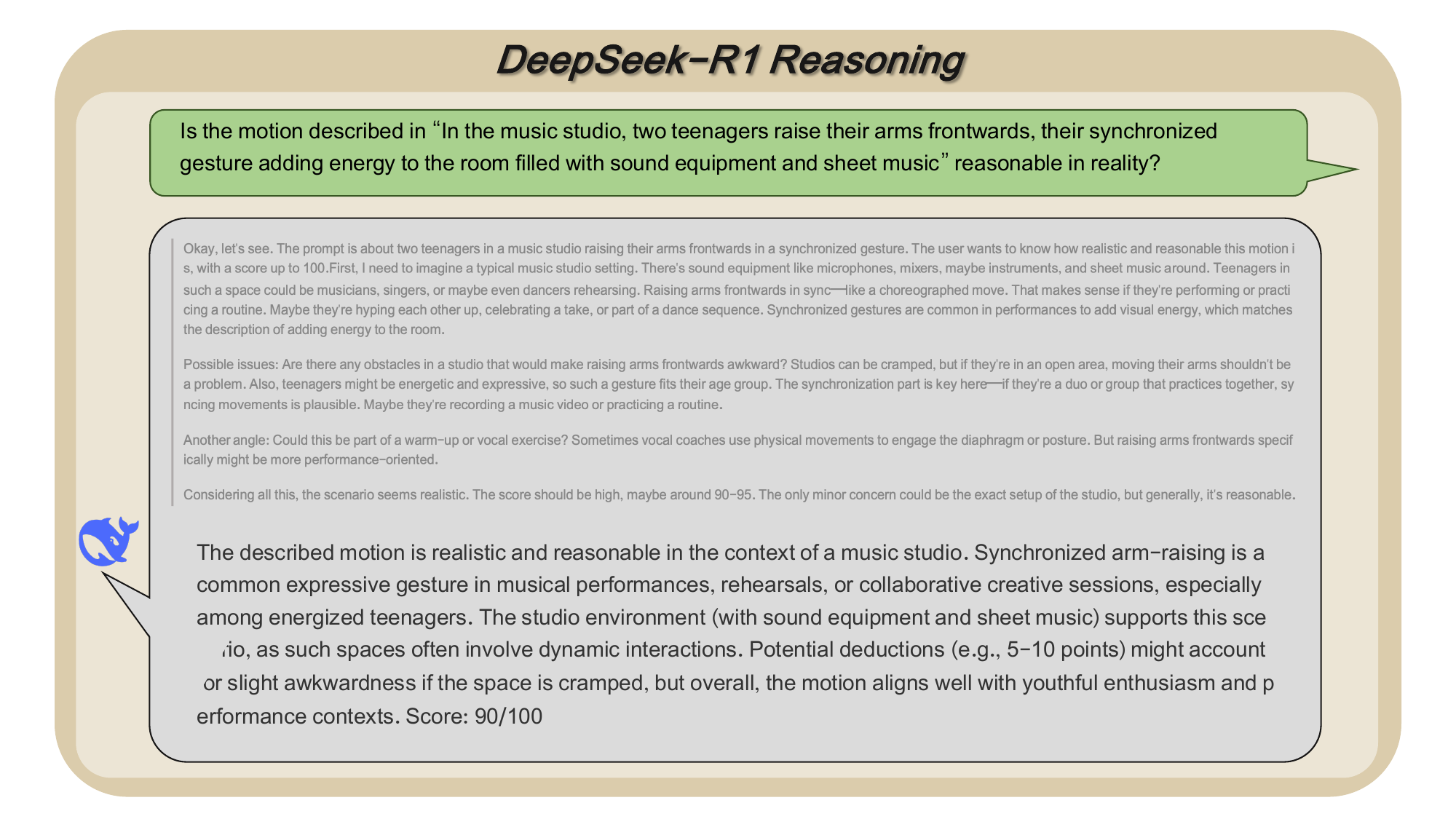}
    \caption{An Example of DeepSeek-R1 Reasoning. A case of evaluating the realistic reasonableness of a prompt using DeepSeek-R1.}
    \label{fig:deepseek}
\end{figure*}

\begin{figure*}
    \centering
    \includegraphics[width=0.9\linewidth]{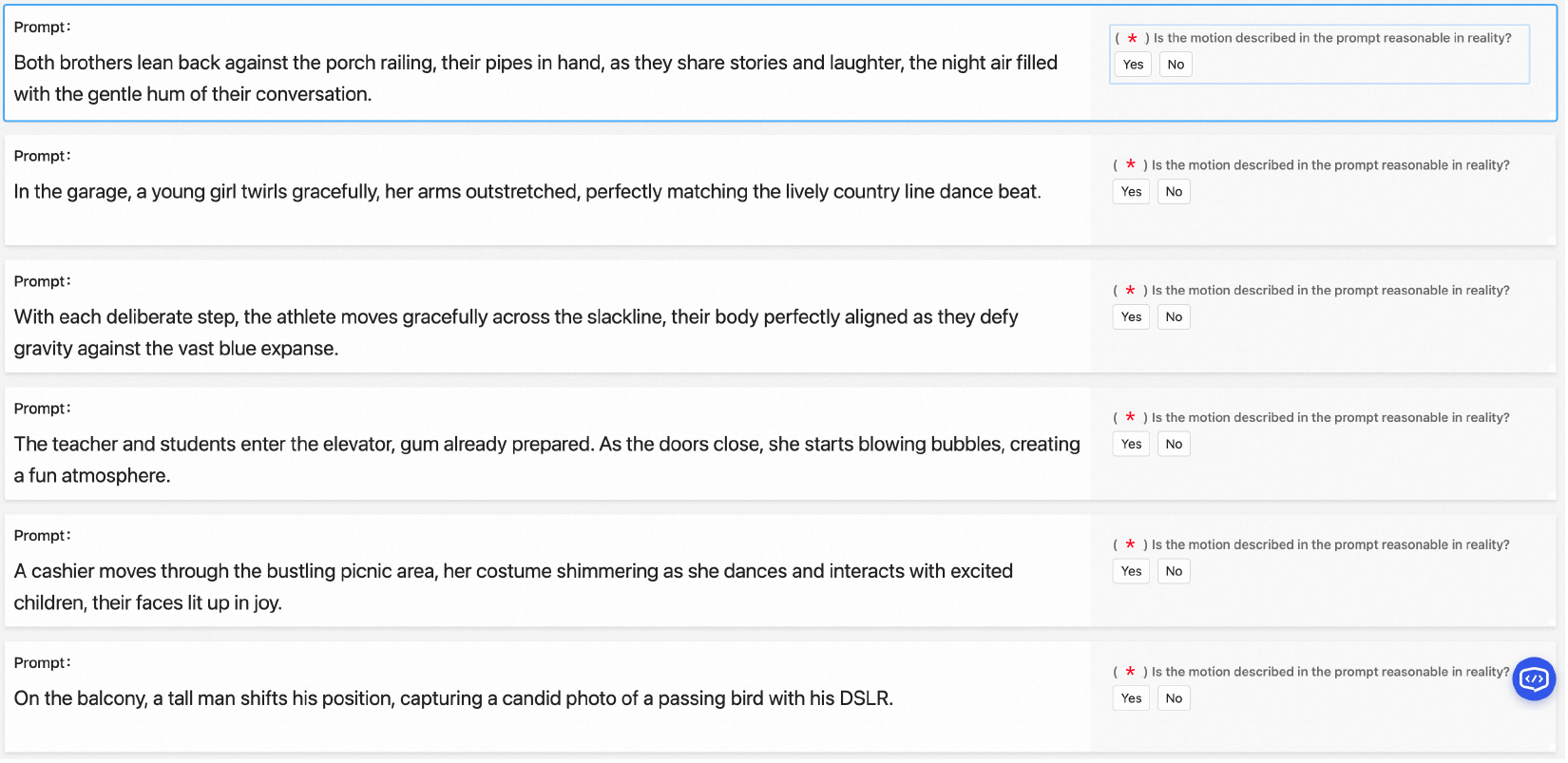}
    \caption{Manual Review of Prompt Validity in Real-World Scenarios. Some cases of manually reviewing the real-world validity of prompts.}
    \label{fig:human review}
\end{figure*}

To ensure that the prompts generated by the GPT-4o describe motion that exists in real life, we combine the efforts of both LLMs and humans to evaluate the plausibility of the prompts. We first utilize the strong reasoning capability of DeepSeek R1~\cite{deepseekai2025deepseekr1incentivizingreasoningcapability} to evaluate the realistic reasonableness of motion descriptions logically~(see Fig.~\ref{fig:deepseek}), which results in a quantified score. After filtering out prompts with lower plausibility scores, we then recruit evaluators to verify the real-world validity of the prompts through a survey~(as shown in Fig.~\ref{fig:human review}). After a rigorous review process, we ultimately retain 1050 prompts that describe reasonably realistic motion.

\begin{figure*}
    \centering
    \includegraphics[width=\linewidth]{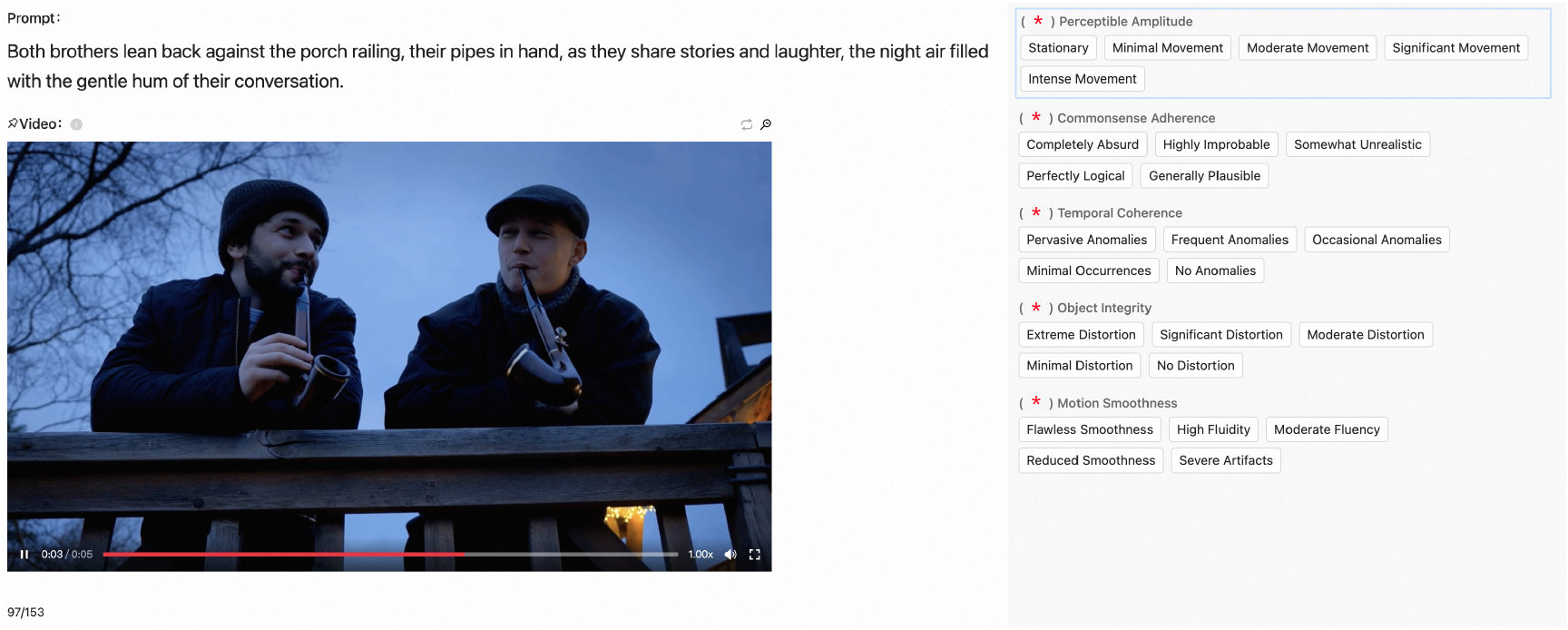}
    \caption{Human Annotation Procedure. Three annotators independently evaluate each aspect, re-watching the video for each question. Annotators are instructed to focus solely on the specific aspect being evaluated, disregarding other potential influences.}
    \label{fig:human-annotation}
\end{figure*}

\section{Implementation Details}

\subsection{Inference Details of Video Generation Models}
\label{inference details}
To ensure a fair comparison, we utilize the best open-source architectures and weights available for each model and maintain the optimal hyperparameters~(including video resolution, sampling steps, scale, etc.) as demonstrated in their respective demos to generate the corresponding videos of approximately 5 seconds. Additionally, we record the time cost of model inference~(excluding model loading) for reference. We list the inference details for each model as follows:

\noindent\textbf{HunyuanVideo}~\cite{kong2024hunyuanvideo} 
The preset video resolution is $624 \times 832$ with a length of 129 frames. Using a 4-GPU parallel inference setup, the generation time for a single video is approximately 610 seconds.

\noindent\textbf{OpenSora}~\cite{zheng2024opensora}
We use the Open-Sora v1.2 model version. The preset video resolution is $720\times1280$ with a length of 102 frames and uses 30 sampling steps. Using a 4-GPU parallel inference setup, the generation time for a single video is approximately 85 seconds.

\noindent\textbf{CogVideoX}~\cite{yang2024cogvideox} 
We use the CogVideoX-5B model version. The preset video resolution is $480\times720$ with a length of 49 frames. Using a 2-GPU parallel inference setup, the generation time for a single video is approximately 355 seconds.

\noindent\textbf{OpenSora-Plan}~\cite{lin2024opensora-plan} 
We use the v1.3.0 model version. The preset video resolution is $352\times640$ with a length of 93 frames. Using a 4-GPU parallel inference setup, the generation time for a single video is approximately 408 seconds.

\noindent\textbf{Mochi 1}~\cite{genmo2024mochi} 
We execute the process with the decode type set to ``tiled full'' and utilize a single GPU pipeline, setting the sampling steps to 64.
The preset video resolution is $480\times848$ with a length of 148 frames. The generation time for a single video is approximately 725 seconds.

\noindent\textbf{Wan2.1}~\cite{wan2.1} 
We use the v1.3.0 model version. The preset video resolution is $352\times640$ with a length of 93 frames. Using a 4-GPU parallel inference setup, the generation time for a single video is approximately 408 seconds.

\subsection{Human Annotation}
\label{sup:annotation}
We recruit three annotators and instruct them to score each video based on five previously defined assessment aspects. These aspects are Commonsense Adherence, Motion Smoothness, Object Integrity Score, Perceptible Amplitude, and Temporal Coherence. For each video's motion quality, the annotators assign scores according to the rating criteria outlined. Our annotation process employs a Likert scale \cite{nemoto2014likert}, with each dimension rated on five levels. Annotators receive detailed descriptions for each dimension to guide their scoring decisions. Our annotation interface is shown in Fig.~\ref{fig:human-annotation}. To ensure a focused evaluation of each aspect, we divide the overall task into five separate annotation packages. In each package, annotators watch the corresponding videos and evaluate only one specific dimension. This approach allows annotators to concentrate on a single aspect of video quality at a time, potentially improving the accuracy and consistency of their assessments. By structuring the annotation process in this way, we aim to obtain more reliable and targeted evaluations for each of the five dimensions of video motion quality.

\section{Qualitative Analysis}
\label{sup:qualitative analysis}
\begin{figure*}[ht]
    \centering
    \includegraphics[width=0.95\linewidth]{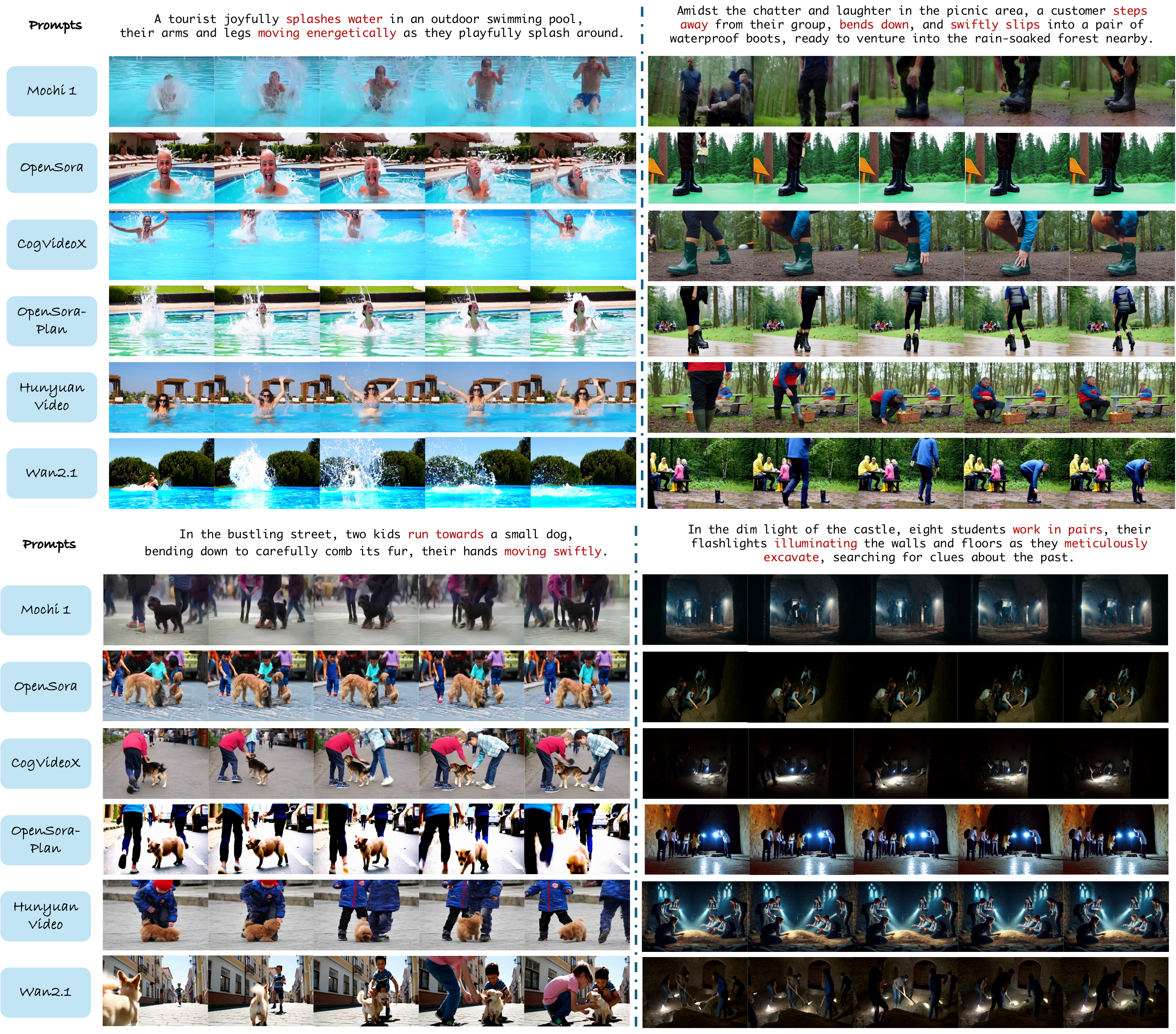}
    \caption{Visualization of Generation Results of Mainstream Models on MMPG-set. Qualitative results on Mochi 1~\cite{genmo2024mochi}, OpenSora~\cite{zheng2024opensora}, CogVideoX~\cite{yang2024cogvideox}, OpenSora-Plan~\cite{lin2024opensora-plan}, HunyuanVideo~\cite{kong2024hunyuanvideo} and Wan2.1~\cite{wan2.1} across six movement modes.}
    \label{fig:qualiative-analysis}
\end{figure*}
To identify where current T2V models exhibit limited capabilities, we qualitatively demonstrate the generation results of T2V models. We select 4 challenging prompts from our benchmark, spanning 6 movement modes for video generation. Fig.~\ref{fig:qualiative-analysis} reveals four critical failure modes: \textbf{Object Persistence Paradox:} Models frequently violate object identity continuity during motion. \textbf{Structural Degeneration:} Dynamic motion induces catastrophic shape distortions. \textbf{Temporal Artifacts:} The generated motion exhibits abrupt discontinuities masked by artificial blurring. \textbf{Newtonian Violations:} Fundamental physics laws are systematically broken, particularly in energy conservation. 

Upon closer examination of the videos generated by various models, we observe significant disparities in quality and adherence to realistic motion. Mochi 1 \cite{genmo2024mochi}, OpenSora \cite{zheng2024opensora}, and OpenSora-Plan \cite{lin2024opensora-plan}, for instance, produce videos plagued by severe blurring and artifacts, substantially degrading overall video quality. While CogVideoX \cite{yang2024cogvideox} and HunyuanVideo \cite{kong2024hunyuanvideo} demonstrate smoother motion, they struggle with maintaining object integrity, often resulting in unnatural distortions of shape during movement sequences.

Notably, we find that Wan2.1 \cite{wan2.1} exhibits the most promising performance among the evaluated models. It generates videos with smooth motion that adhere well to basic physical principles, aligning closely with our fundamental visual expectations. Upon careful observation of task-specific details such as object shapes and limb movements, Wan2.1's outputs appear more natural and consistent. Moreover, it demonstrates a superior ability to accurately represent the amplitude and scale of specific movements as described in the prompts.

These observations underscore the ongoing challenges in text-to-video generation, particularly in maintaining consistency, physical plausibility, and natural motion across diverse scenarios. While progress is evident in some models, there remains significant room for improvement in addressing these critical aspects of video generation.

\end{document}